\pdfoutput=1
\documentclass[11pt]{article}
\usepackage[utf8]{inputenc}
\usepackage[T1]{fontenc}
\usepackage{lmodern}
\usepackage[margin=1in]{geometry}
\usepackage{graphicx}
\usepackage{booktabs}
\usepackage{array}
\usepackage{amsmath}
\usepackage{tikz}
\usetikzlibrary{arrows.meta,positioning,shapes.misc,calc}
\usepackage[numbers,sort&compress]{natbib}
\usepackage{caption}
\usepackage{subcaption}
\usepackage{microtype}
\usepackage{xcolor}
\usepackage[colorlinks=true,linkcolor=blue!50!black,citecolor=blue!50!black,urlcolor=blue!50!black]{hyperref}
\title{Where Should Language Sit in a Multimodal Model?\\Lessons from What Language Does\\to Human Perception and Cognition}
\author{Peng Xie\thanks{Peng Xie and Amr Alanwar are with the TUM School of Computation, Information and Technology, Department of Computer Engineering, Technical University of Munich.}\\\texttt{p.xie@tum.de} \and Amr Alanwar\footnotemark[1]\\\texttt{alanwar@tum.de}}
\date{}

\begin{document}
\maketitle

\begin{abstract}
Language models compute over tokens: language is their input, their output, and increasingly their internal representation. Whether language should keep all of these positions depends on what language does to the system that uses it. The one system with a century of data on that question is the human. We review what language does to human perception, the brain, and thought, and read the same evidence against multimodal models and language models. Throughout, we treat language as a compressor that runs on a shared codebook: a word is an index, the content is in the receiver, and a community maintains the codebook. In humans the compression is measurable, learning the codebook reorganizes the senses, and thought survives the loss of language. We then measure the rule that models apply when two cues disagree, with cue-conflict experiments on six vision-language models and two robot policies. Surviving cues are weighted in the order their reliabilities prescribe, at 11 to 82\% of the ideal observer's slope, and many answers copy the text. One policy family drops a cue that adds no information beyond the others rather than down-weighting it, another keeps it at a weight that fails when the cues conflict, and a visual cue that identifies the task in every training frame is never learned, because the language pathway already fits the data. Language models are the best current models of the human language network, and they have entered the human speech community, shifting word frequencies while alignment narrows their conceptual diversity. We close with seven implications for token-based systems. Language belongs at a model's boundary and in the shared codebook, as in the brain, not as its internal representation; the price of leaving the codebook inside is auditability.
\end{abstract}

\section{Introduction}
\label{sec:intro}

In early 2025 a hackathon demo circulated online. Two voice agents were talking to each other in English. A few sentences in, each confirmed that the other was also an AI, and the two dropped English for a modem-like acoustic protocol that no human listener could follow \citep{gibberlink2025}. The choice was reasonable. Across 17 languages, human speech transmits about 39 bits per second whatever the speaking rate \citep{coupe2019}. Reading runs at 28 to 45 bits per second, and the whole of human behavioral output amounts to about 10 bits per second, against a sensory input of about $10^9$ \citep{zheng2024}. The rate of language is matched to the serial stage of human cognition, not to the senses. A machine has no such limit, and two agents have no reason to keep talking in a code designed for a listener who takes in 10 bits per second.

Yet the language models we build put language in every position. A transformer reads tokens and predicts the next one \citep{vaswani2017}, so language is its input and its output. Chain-of-thought prompting makes it reason in language as well \citep{wei2022}. Most multimodal systems convert images, audio, and robot actions into tokens before a language backbone processes them \citep{liu2023llava,borsos2023,brohan2023}. Should language keep all four positions, as input, as output, as internal representation, and as the channel between agents? Comparing architectures on benchmarks cannot settle this, because the answer changes with the architecture and the training data. Settling it requires knowing what language is for and what it does to the system that uses it. Whether language is the source of general intelligence or only an interface to it is argued from both sides \citep{lupyan2026important,tenenbaum2026language}. The one system with more than a century of data on that question is the human.

To the senses, language is a compressor. The idea has a long history: Zipf's principle of least effort \citep{zipf1949}, word lengths tuned to information content \citep{piantadosi2011}, word meanings that support efficient communication \citep{regier2015}, and, most directly, the finding that lexicons across languages sit near the optimum of a compression objective \citep{kirby2015,zaslavsky2018,gibson2019}. Say ``car'' and the listener pictures a box on four wheels. Without the word, describing the object takes a long string of sentences. With the word, a few bytes suffice, where a photograph of a car costs a few hundred thousand. The compression ratio is enormous, and the compression is lossy: the car in your head is not the car in mine \citep{marti2023}. A parlor game makes this visible. One person looks at a picture and describes it to a second person who cannot see it, and the second person draws it. The drawing is never identical to the original, but it is close enough that a third person can guess what the picture shows. The game now has an engineering version: an image is compressed to a caption plus a little side information, and a generative model draws it back, at rates down to a few thousandths of a bit per pixel \citep{lei2023,careil2024} (Section~\ref{sec:vision}). Language compresses visual information about as far as it can go, and what it throws away is exactly the part that differs from one head to the next.

The compression works not because the word ``car'' contains a car, but because speaker and listener hold the same lookup table. We call that table a codebook, the term in information theory for a table of codes that sender and receiver share. A word is an index into the codebook. The content, the box on wheels, is in the table, and the table is something a community learns from one another and maintains together. The view is old. Saussure stated the arbitrariness of the sign in 1916, Wittgenstein the privacy of content in 1953, and Putnam the division of linguistic labor in 1975 \citep{saussure1916,wittgenstein1953,putnam1975}. We use it here to put evidence from several disciplines side by side, and because it applies unchanged to a language model. A model has no senses, yet it learns the whole codebook from text, much as a person born blind learns the appearance of animals from language, with color the dimension learned least well \citep{kim2019,lewis2019}. What such a model can and cannot get from text is the subject of a long debate in natural language processing (Section~\ref{sec:grounding}). And models have begun to take part in maintaining the codebook \citep{yakura2024,murthy2025,brinkmann2023machine}.

If language is a compressor that runs on a shared codebook, three things should follow, and each can be tested on humans. First, the compression should be measurable. Color vocabularies in more than a hundred languages lie near the information-theoretic optimum \citep{zaslavsky2018}, and describing a face in words makes it harder to recognize afterwards, an effect that replicates, at a size that depends on the timing of the description \citep{schooler1990,alogna2014}. Second, the codebook has to be learned, and learning it should alter the senses that feed it. Phonetic contrasts that infants can hear at six months are gone by twelve \citep{werker1984,kuhl2006}. Russian has two words for blue, Russian speakers discriminate those two blues faster, and the advantage disappears under a concurrent verbal task \citep{winawer2007}. Learning to read reorganizes visual cortex, with effects that reach back to V1 \citep{dehaene2010}. Third, the codebook is a protocol for communication rather than the place where thinking happens, so thought should survive the loss of language, and in severe aphasia it does: arithmetic, algebra, and reasoning about other minds remain \citep{fedorenko2024,varley2005,klessinger2007,varley2000}.

The second half of the paper asks three parallel questions of models: whether learning the codebook shapes what they take from their senses, whether representation and reasoning survive without it, and what happens when they join the community that maintains it. Vision-language models fail simple visual questions and often answer without using the image \citep{tong2024,rahmanzadehgervi2024,zhou2026}, audio language models answer questions about a voice from its transcript \citep{pang2026voxparadox}, and vision-language-action models act without attending to the instruction \citep{lian2026,fei2025}. We measure the rule behind both with the cue-conflict method of psychophysics (Section~\ref{sec:cueconflict}). The models re-weight image and text in the direction that reliability prescribes, though short of the ideal observer. A cue that adds no information in the task can be dropped entirely, where the brain's rule would keep it at a weight set by its reliability \citep{ernst2002} and discard it only on inferring a separate source \citep{kording2007}. The same architecture completes the instructed task on 97\% of trials in a scene where only the instruction identifies the task, and on 0\% in scenes where the layout identifies it; a single policy trained on both kinds of scene gives 98\% and 0\%. A second policy family, $\pi_{0.5}$, obeys the swapped instruction on 28\% of the same episodes and completes neither task on 42\%, so how far a redundant cue is dropped depends on the training recipe, and a task-predictive visual tag added in fine-tuning is not learned at all.

Visual representations trained with no language at all now match language-supervised ones \citep{fan2025}, and models that reason in a continuous latent space or in images beat verbal chains of thought on several tasks \citep{hao2024,xu2025}. Language models are the best current models of the human language network \citep{schrimpf2021}, and that relation is already used in both directions: fMRI signals improve model reasoning \citep{xiao2026}, and models decode continuous language from fMRI \citep{tang2023}. Finally, language models have joined the community that maintains the codebook. Words that ChatGPT favors rose abruptly in spontaneous human speech within two years of its release \citep{yakura2024}, and alignment reduces the conceptual diversity of the models themselves, none of which reaches the diversity of a human population \citep{murthy2025}.

Sections~\ref{sec:codebook} to~\ref{sec:technology} present the human evidence. Section~\ref{sec:codebook} sets out the codebook, and Section~\ref{sec:bandwidth} the information rates of the senses and the rule by which the brain combines them. Section~\ref{sec:perception} covers the effects of language on perception, and Section~\ref{sec:technology} language as a cognitive technology, its separation from thought, and the evidence from animals. Sections~\ref{sec:models} and~\ref{sec:exit} present the models. Section~\ref{sec:claims} draws seven implications for token-based systems and lists the open problems.

\section{Language as a community-maintained codebook}
\label{sec:codebook}

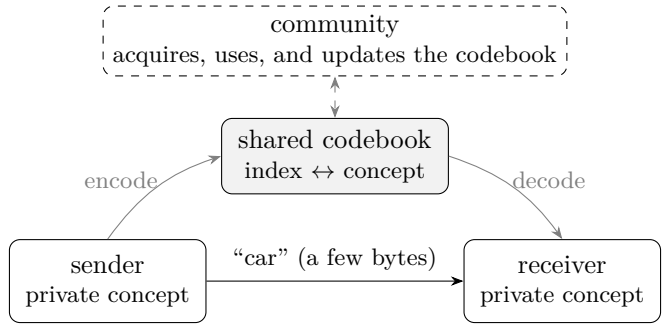
\begin{figure}[t]
\centering
\begin{tikzpicture}[>=Stealth, node distance=8mm, every node/.style={font=\small}]
\node[draw, rounded corners, minimum width=2.6cm, minimum height=1.1cm, align=center] (spk) {sender\\[-1pt]\footnotesize private concept};
\node[draw, rounded corners, minimum width=2.6cm, minimum height=1.1cm, align=center, right=3.4cm of spk] (lis) {receiver\\[-1pt]\footnotesize private concept};
\node[draw, fill=gray!10, rounded corners, minimum width=3.0cm, minimum height=1.0cm, align=center, above=1.15cm of $(spk)!0.5!(lis)$] (cb) {shared codebook\\[-1pt]\footnotesize index $\leftrightarrow$ concept};
\node[draw, dashed, rounded corners, minimum width=4.0cm, minimum height=0.8cm, align=center, above=0.55cm of cb] (com) {community\\[-1pt]\footnotesize acquires, uses, and updates the codebook};
\draw[->] (spk) -- node[above, font=\footnotesize] {``car'' (a few bytes)} (lis);
\draw[->, gray] (spk.north) to[bend left=18] node[left, font=\footnotesize, pos=0.6] {encode} (cb.west);
\draw[->, gray] (cb.east) to[bend left=18] node[right, font=\footnotesize, pos=0.4] {decode} (lis.north);
\draw[<->, gray, dashed] (com) -- (cb);
\end{tikzpicture}
\caption{The organizing description. A word is an index into a codebook that sender and receiver share. The community builds and maintains the codebook. The concepts at the two ends are private and differ (Wittgenstein's beetle-in-the-box; \citealp{marti2023}).}
\label{fig:codebook}
\end{figure}

This section gives the evidence that the codebook description (Figure~\ref{fig:codebook}) is more than a metaphor. The entries in the table are private and differ from person to person (Section~\ref{sec:private}). The table is shaped by pressures toward efficiency and is shared down to the level of brain activity (Section~\ref{sec:emergence}). It is costly to acquire, and at community scale it is found in no other species (Section~\ref{sec:acquisition}). The description also takes a position in the debate about whether a model trained on text alone can have meaning (Section~\ref{sec:grounding}).

\subsection{Arbitrariness, private content, and the division of linguistic labor}
\label{sec:private}

Philosophers made the basic points between 1916 and 1980. The link between a word and its object is arbitrary \citep{saussure1916}. An apple has a color, a hardness, and a caloric content that an instrument can measure; it has no property called ``apple''. Wittgenstein's beetle-in-the-box argument separates what a word does in public from what it contains in private. Imagine that everyone owns a box that only they can look into, and everyone calls whatever is inside a beetle. The word ``beetle'' works in conversation whether or not the boxes hold the same thing \citep[\S293]{wittgenstein1953}. Kripke described how a name gets attached to its object: someone names it once, and a chain of speakers passes the name on \citep{kripke1980}. Putnam added the division of linguistic labor. A speaker who cannot tell an elm from a beech still uses ``elm'' correctly, because the community keeps experts who can \citep{putnam1975}.

Each of these points now has data behind it. For common nouns such as ``penguin'', clustering finds 5 to 13 measurably different meanings among 100 people, and the authors estimate 10 to 30 variants for a common noun in the population; speakers do not know this: each assumes that others share their own variant \citep{marti2023}. That is the beetle-in-the-box measured, and it explains part of why people talk past one another. Nicaraguan Sign Language shows the need for a community directly. Deaf children at a newly expanded school turned a set of home-sign systems into a language with hierarchical structure within two cohorts \citep{senghas2004}, and the innovations came from signers who had entered the community as young children, before about age ten \citep{senghas2001}. Adults who acquired only the early form of the language as children also do worse on false-belief tasks \citep{pyers2009}. Symbols need interaction to form. In graphical communication experiments, drawings turn into arbitrary marks only when partners exchange feedback, and people outside the pair cannot read the result \citep{garrod2007,fay2018}. Descriptions shrink in the same way: the same tangram figure goes from ``a person who's ice skating, except they're sticking two arms out in front'' to ``the ice skater'' over repeated rounds as the pair builds common ground \citep{clark1986}.

\subsection{Emergence, efficiency, and neural coupling}
\label{sec:emergence}

Two pressures shape what goes into the codebook. Iterated learning experiments pass an artificial language along a chain of learners. Without anyone designing it, the language becomes more learnable and more structured \citep{kirby2008}, and compositional structure appears only when a pressure to be compressible meets a pressure to stay expressive \citep{kirby2015}. The same trade-off predicts the vocabulary. Color naming systems across languages lie near the information-bottleneck optimum \citep{zaslavsky2018}, and color terms elicited from language models sit near the same frontier \citep{imel2026evolution}; so do names for containers and animals \citep{zaslavsky2019}, personal pronoun systems \citep{zaslavsky2021}, and grammatical markers of number, tense, and evidentiality \citep{mollica2021}. Kinship terminologies are near-optimal on the related trade-off between simplicity and informativeness \citep{kemp2012}. \citet{gibson2019} review the broader case that communicative efficiency shapes language. Language models apply a harder compression than people do: their embeddings recover human category boundaries but discard fine distinctions that people keep \citep{shani2026tokens}.

The sharing can be seen in the brain. When someone tells a story, the listener's cortical activity becomes coupled to the speaker's, and the coupling disappears when communication fails \citep{stephens2010}. Contextual embeddings from a language model can track a word from the speaker's brain, where it appears before it is spoken, to the listener's brain, where it appears after \citep{zada2024}. Speakers of English, Chinese, and French who hear the same story show activity in one common conceptual space, and models trained on the three languages land in the same space \citep{zada2025}. Language models converge in the same way internally: their middle layers place semantically equivalent inputs from different languages and modalities near one another \citep{wu2025}, and larger vision and language models measure distances between data points in increasingly similar ways \citep{huh2024}.

Arbitrariness is not total. Across nearly two thirds of the world's languages, the word for ``tongue'' tends to contain /l/ and the word for ``small'' tends to contain /i/ \citep{blasi2016}. In the bouba/kiki test, about 72\% of responses across 25 languages and 10 writing systems match the expected shape, and the effect is reliable in 17 of the 25 languages \citep{cwiek2022}. Sound symbolism constrains names slightly. It does not supply them.

\subsection{Acquisition cost and the comparative record}
\label{sec:acquisition}

Learning the codebook is expensive. A 20-year-old native speaker of American English knows about 42,000 lemmas \citep{brysbaert2016} and reads non-fiction at 238 words per minute \citep{brysbaert2019}. By age 20 a person has taken in somewhere between 30 million and 400 million words, three to five orders of magnitude less language than a large language model is trained on \citep{frank2023}. The economics are those of a shared standard with network externalities \citep{katz1985}. Each member pays the fixed cost of learning once, each message afterwards costs almost nothing, and the larger the community, the lower the cost of maintenance per member.

Whether the codebook is unique to humans depends on which feature one counts. Hockett's design features \citep{hockett1960} and the distinction between the broad and narrow faculties of language \citep{hauser2002} reserve recursion and open-ended productivity for humans, and \citet{deacon1997} puts the break at symbolic reference. Field studies keep finding isolated entries in other species. African elephants address individuals with calls that do not imitate the receiver \citep{pardo2024}, marmosets label one another with phee calls \citep{oren2024}, and bottlenose dolphins address each other by copying signature whistles \citep{king2013}. Vervet monkeys give different alarm calls for leopards, eagles, and snakes, and listeners respond to playback by climbing, looking up, or scanning the ground \citep{seyfarth1980}. Japanese tits combine calls in a fixed order, and reversing the order removes the response \citep{suzuki2016}. Sperm whale codas have combinatorial structure \citep{sharma2024}. Vocal learning itself is rare. It evolved independently in songbirds, parrots, and hummingbirds among birds and in humans, cetaceans, pinnipeds, bats, and elephants among mammals, and it is at most rudimentary in other primates \citep{jarvis2019}, which is why a bonobo answers with a keyboard rather than a voice (Section~\ref{sec:animals}). Artificial agents that have to coordinate also invent shared codes \citep{lazaridou2017,mordatch2018}. What no other species has built is a codebook with tens of thousands of entries, maintained across generations by a community, and combined by rule.

\subsection{Form, meaning, and grounding in language models}
\label{sec:grounding}

The codebook description takes a side in a debate that cognitive science and natural language processing have conducted for more than thirty years. \citet{harnad1990} posed the symbol grounding problem: a system whose symbols are defined only by other symbols is like a learner working from a Chinese-to-Chinese dictionary, and symbols must be grounded bottom-up in sensory representations. \citet{benderkoller2020} restated the problem for language models. A system trained only on the form of language has, a priori, no way to learn meaning. \citet{merrill2021} gave the formal version: the assertions in ungrounded text let a learner emulate meaning for languages with a strong form of semantic transparency, and once the same expression can take different values in different contexts, emulation can become uncomputable. The other side holds that distributional structure carries much of meaning. \citet{piantadosihill2022} argue from conceptual-role semantics that meaning arises from how a system's internal states relate to one another, so a model trained on text can capture important aspects of it. \citet{sogaard2023} argues that reference itself is probably learnable from raw text, because higher-order co-occurrence statistics are stable across languages and modalities. \citet{coelhomollo2026} argue that language models can meet the conditions for referential grounding without a body, and \citet{pavlick2023} holds that the objections from missing symbols and missing grounding are premature until the models' representations have been characterized. A third position insists on experience. \citet{bisk2020} argue that shared experience of the world is what makes utterances meaningful, and \citet{lakemurphy2023} argue that distributional models approximate human judgments of word similarity but not the perceptual content, goals, and beliefs that words carry. \citet{mitchell2023} survey the debate.

The codebook description sorts these claims by what text contains. Text contains the indices and the relations among them, so a model trained on text can recover the structure of the codebook: which entries are similar, where they sit in a conceptual space, and how they combine (Section~\ref{sec:emergence}). \citet{taniguchi2026generative} describe the same object from the model side: a language model approximates the collective world model that a community's communication encodes. Text does not contain the content of the entries. That content sits in the receiver, and the senses put it there, so a model without sensors holds entries whose content is inferred from their neighbors rather than perceived. The human parallel is the person born blind, whose knowledge of animal appearance, learned from language, agrees with sighted knowledge on most dimensions and diverges on color (Section~\ref{sec:vision}). The same distinction says what language is for and what it costs. The senses deliver high-dimensional streams that no two people can compare directly; the codebook is the one space in which they can be pooled and exchanged, because it is discrete and shared. The price is that the space holds indices, so whatever a system cannot look up in its own perception is lost, and Section~\ref{sec:ignore} shows that price paid by models in three modalities. Section~\ref{sec:claims} returns to what this implies for grounding.

\section{Information rates from sensory input to behavioral output}
\label{sec:bandwidth}

This section asks three quantitative questions: how much information the senses deliver, how much of it the brain can act on, and how the brain combines the senses when they disagree. The first two answers set the budget that language operates under; the third is the rule against which Section~\ref{sec:cueconflict} measures multimodal models.

\begin{figure}[t]
\centering
\includegraphics[width=\linewidth]{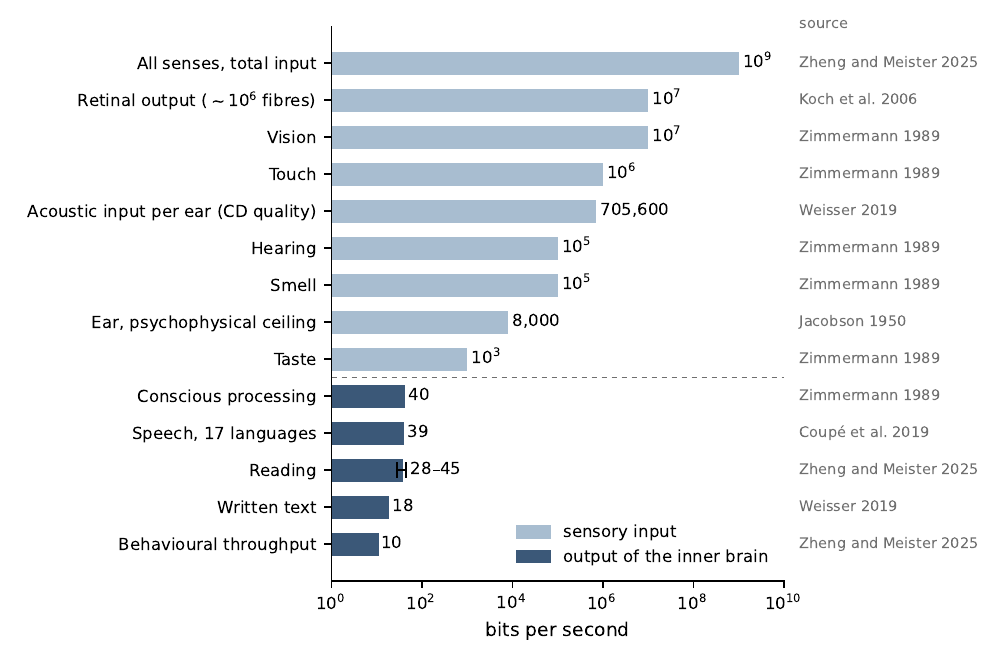}
\caption{Information rates along the path from the senses to behavior, in bits per second on a log scale. Light bars are sensory input; dark bars are outputs of the slow serial stage that \citet{zheng2024} call the inner brain. Horizontal lines mark reported ranges. Sources: \citet{zheng2024,koch2006,zimmermann1989,weisser2019,jacobson1950,coupe2019}. The Zimmermann figures are estimates from the 1980s, and no modern aggregate estimate exists for touch.}
\label{fig:bandwidth}
\end{figure}

\subsection{Sensory input rates and the capacity of behavior}

Figure~\ref{fig:bandwidth} collects the estimates. They are accurate to about an order of magnitude, which is enough for the argument. On the input side, the retina sends about $10^6$ output fibers to the brain \citep{koch2006}. Recordings from seven types of ganglion cell in the guinea pig give about 875,000 bits per second for that retina. Scaling to the $10^6$ ganglion cells of the human retina gives roughly $10^7$ bits per second, which the authors compare to an Ethernet connection \citep{koch2006}. Zimmermann's table, popularized by N{\o}rretranders, puts vision at $10^7$ bits per second, touch at $10^6$, hearing and smell at $10^5$, and taste at $10^3$, against a maximum of about 40 bits per second of conscious processing \citep{zimmermann1989,norretranders1998}. Estimates of this kind go back to Jacobson's work of the 1950s, which put the ear at about 8,000 bits per second and a single cochlear fiber at about 0.3 bits per second \citep{jacobson1950}.

Modern measurements exist for single fibers but not for whole channels. The auditory nerve has about 31,000 myelinated fibers \citep{spoendlin1989}, and a single fiber carries two to six times more information per second for natural sounds than for broadband noise, close to the limit set by its spike statistics \citep{rieke1995}. The glabrous skin of one hand has about 17,000 mechanoreceptive units \citep{johansson1979}. A single afferent conveys between a fraction of a bit and about one bit about force direction or surface curvature per contact, much of it already in the timing of its first spikes \citep{saal2009}. No modern estimate exists for the tactile channel as a whole, so Zimmermann's $10^6$ is still the only aggregate figure. \citet{weisser2019} traces the auditory chain from the outside in. An ear receives 705,600 bits per second at CD quality, and telephony compresses that to 4.75 to 23.85 kilobits per second while speech stays intelligible. Cognition handles about 52 bits per second, speech carries about 35, and text about 18.

The output side is far narrower. \citet{zheng2024} put total sensory input near $10^9$ bits per second and total behavioral output near 10, a gap of eight orders of magnitude. They describe an outer brain that runs at the high rate and an inner brain that runs at the low one. Speech transmits about 39 bits per second across 17 languages regardless of speaking rate, because faster languages pack fewer bits into each syllable \citep{coupe2019}. Reading runs at 28 to 45 bits per second, and comprehension suffers above that \citep{zheng2024}. LeCun's arithmetic reaches the same conclusion from the input side: language delivers under 12 bytes per second, while a four-year-old child has taken in about $10^{15}$ bytes through the eyes, fifty times the text in the largest training sets \citep{lecun2024}.

The 10 bits per second figure is contested. \citet{sauerbrei2025} object, in a Comment, that the limit appears to hold for deliberate cognition but not for the unconscious control of movement, which on their account engages most of the neurons in the central nervous system and carries most of a human's information throughput. The argument here uses the figure only for language, reading, and deliberate choice, the high-level functions that the limit does describe and the functions that a language model imitates.

The limit on language is therefore set downstream of the senses, not by them. The whole conscious stream amounts to at most a few tens of bits per second \citep{zimmermann1989,weisser2019,zheng2024}, and language re-encodes that stream into a form that can leave the body and be shared. \citet{christiansen2016} draw the consequence for the listener: because memory for the incoming signal is fleeting, language has to be compressed and recoded as it arrives, chunk by chunk, a constraint they call the Now-or-Never bottleneck.

Two facts about humans explain two different features of language, and the two should be kept apart. The first is that minds are separate. No memory is shared between two brains, and the only path between them runs through motor output and sensory input over a noisy channel. That fact explains why a shared, discrete codebook exists at all. A discrete combinatorial code overcomes the limit that noise places on the number of distinguishable signals \citep{hockett1960,nowak1999}, and fast brains would need it as much as slow ones. The second fact is that the serial stage of human cognition runs at tens of bits per second \citep{zheng2024,coupe2019}. That fact explains the parameters of the code, its rate and its compression ratio \citep{christiansen2016,kirby2015,zaslavsky2018}. The parameters are a compromise between rate and distortion made for a human reader. The codebook is not. A model that talks to people inherits the first fact and can discard the second.

\subsection{Cortical allocation and metabolic cost}

The allocation of cortex tells the same story. In the macaque about three quarters of the cortical surface serves the senses and movement, with visual cortex alone at 52\%; in humans the share is about half, visual cortex takes 27\%, and V1 covers 2.2\% of cortex, one sixth of its share in the macaque \citep{felleman1991,vanessen2004}. Across species the layout of cortical fields follows the sensory niche \citep{krubitzer2007}, and within a sense neurons are allocated in proportion to how probable each stimulus is \citep{ganguli2014}, a constraint that predicts a set of perceptual biases that were otherwise unexplained \citep{wei2015}. Rate-distortion theory, which minimizes error under a fixed bit budget, is the same mathematics as the information bottleneck \citep{tishby1999} behind color naming \citep{zaslavsky2018}; it accounts for perceptual and memory data \citep{sims2016,bates2020} and derives the universal law of generalization \citep{sims2018}. The compression is forced by energy cost. Communication between neurons uses 35 times the energy of computation in human cortex \citep{levy2021}, and action potentials and synaptic transmission take most of the grey-matter energy budget \citep{attwell2001}. Axon volume and energy rise with the square of transmission rate, so the design rule is to send only what is needed, as slowly as possible \citep{sterling2015}. Attention and awareness are measurable bottlenecks \citep{marois2005}, and global workspace theory treats consciousness itself as a broadcast channel of limited capacity \citep{baars1988,dehaene1998,dehaene2001,mashour2020}.

\subsection{Reliability-weighted multisensory integration}
\label{sec:fusion}

When the senses disagree, the brain trusts the one that is currently more reliable. Vision and touch combine like a maximum-likelihood estimator: touch takes over when visual noise is added \citep{ernst2002}. The ventriloquist effect is the same weighting applied to sight and sound \citep{alais2004}. One flash paired with two beeps is seen as two flashes \citep{shams2000}. The brain also weighs how likely two signals are to share a source and combines them in proportion to that belief, a computation known as Bayesian causal inference \citep{kording2007}. The general rule is called modality appropriateness: the more reliable sense for a given dimension wins, vision for space and hearing for time \citep{welch1980}. The weights change with age: visual dominance in simple detection tasks is present in adults and absent in children \citep{hirst2018}. The stream can even be re-routed. Blind subjects learned to see through a camera whose image was projected onto the skin of their back \citep{bachyrita1969}, and in people born blind, occipital cortex processes sentences \citep{bedny2011}. Integration ends in an amodal hub in anterior temporal cortex, while the content stays in the sensory and motor areas that feed it, the hub-and-spoke arrangement of \citet{patterson2007}, and along the way concept representations are abstracted across modalities in a hierarchy of multimodal areas \citep{fernandino2016}. Words attach at the hub, not at the spokes. Section~\ref{sec:cueconflict} measures whether multimodal models apply this rule.

\section{Effects of language on perception}
\label{sec:perception}

What language does to perception differs by sense. Vision has the most evidence, from the compression of pictures into signs to the effect of labels on perception and on cortex (Section~\ref{sec:vision}). Audition shows the ear being tuned to the native language and shows what the voice carries that text drops (Section~\ref{sec:hearing}). The remaining modalities show that the language determines which senses can be put into words (Section~\ref{sec:other}).

\subsection{Vision: lexical encoding, verbal overshadowing, and categorical perception}
\label{sec:vision}

\begin{figure}[!tp]
\centering
\begin{subfigure}[t]{\linewidth}
\centering
\includegraphics[width=0.9\linewidth]{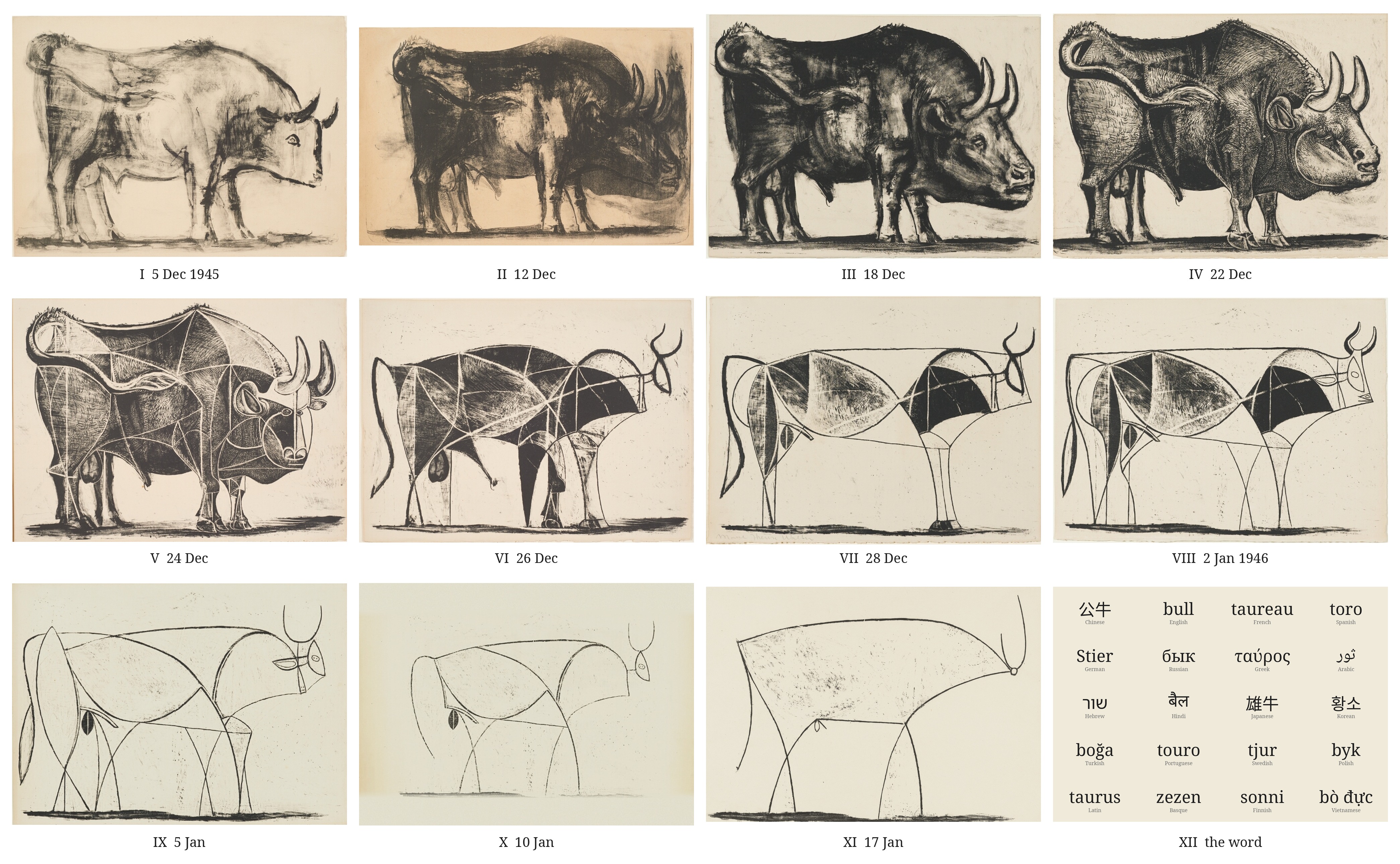}
\caption{Picasso, \emph{Le Taureau}, eleven lithographic states from 5 December 1945 to 17 January 1946 (states I--X: Norton Simon Museum, M.1977.08.3.01--10; state XI: Museum of Modern Art; reproduced for scholarly commentary), followed by a twelfth panel that the artist never drew: the word for the animal in twenty languages.}
\end{subfigure}\\[6pt]
\begin{subfigure}[t]{\linewidth}
\centering
\setlength{\tabcolsep}{6pt}
\begin{tabular}{@{}c c c c c c c c c@{}}
\includegraphics[height=1.25cm]{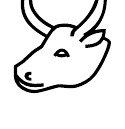} & $\rightarrow$ &
\includegraphics[height=1.25cm]{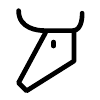} & $\rightarrow$ &
\includegraphics[height=1.25cm]{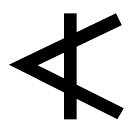} & $\rightarrow$ &
\includegraphics[height=1.25cm]{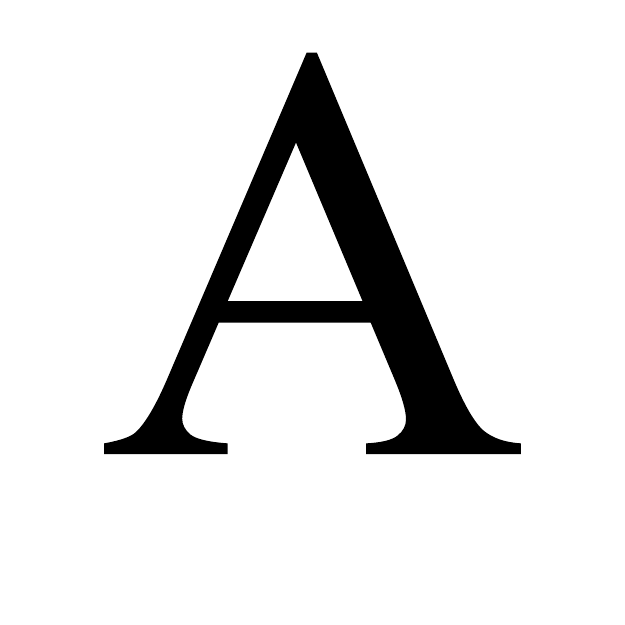} & $\rightarrow$ &
{\fontsize{34}{40}\selectfont A}\\
\footnotesize ox head (Egyptian F1) & & \footnotesize Proto-Sinaitic & & \footnotesize Phoenician \emph{aleph} & & \footnotesize Greek alpha & & \footnotesize Latin
\end{tabular}
\caption{The letter A. An ox-head pictogram became the Proto-Sinaitic sign for \emph{'alp} ``ox'', the Phoenician \emph{aleph}, the Greek alpha, and the Latin A \citep[][secs.~5, 21, and 24]{goldwasser2010,daniels1996}.}
\end{subfigure}\\[6pt]
\begin{subfigure}[t]{\linewidth}
\centering
\setlength{\tabcolsep}{6pt}
\begin{tabular}{@{}l c c c c c c c@{}}
\emph{niu} ``ox'' &
\includegraphics[height=1.15cm]{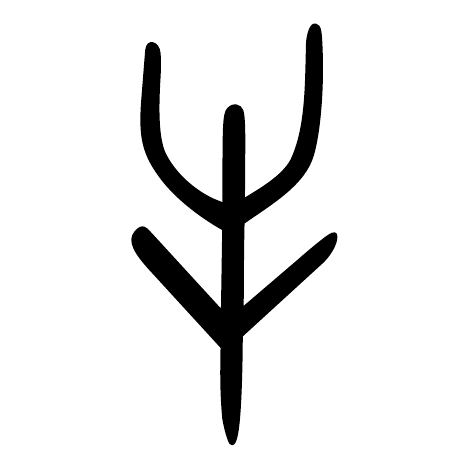} & $\rightarrow$ &
\includegraphics[height=1.15cm]{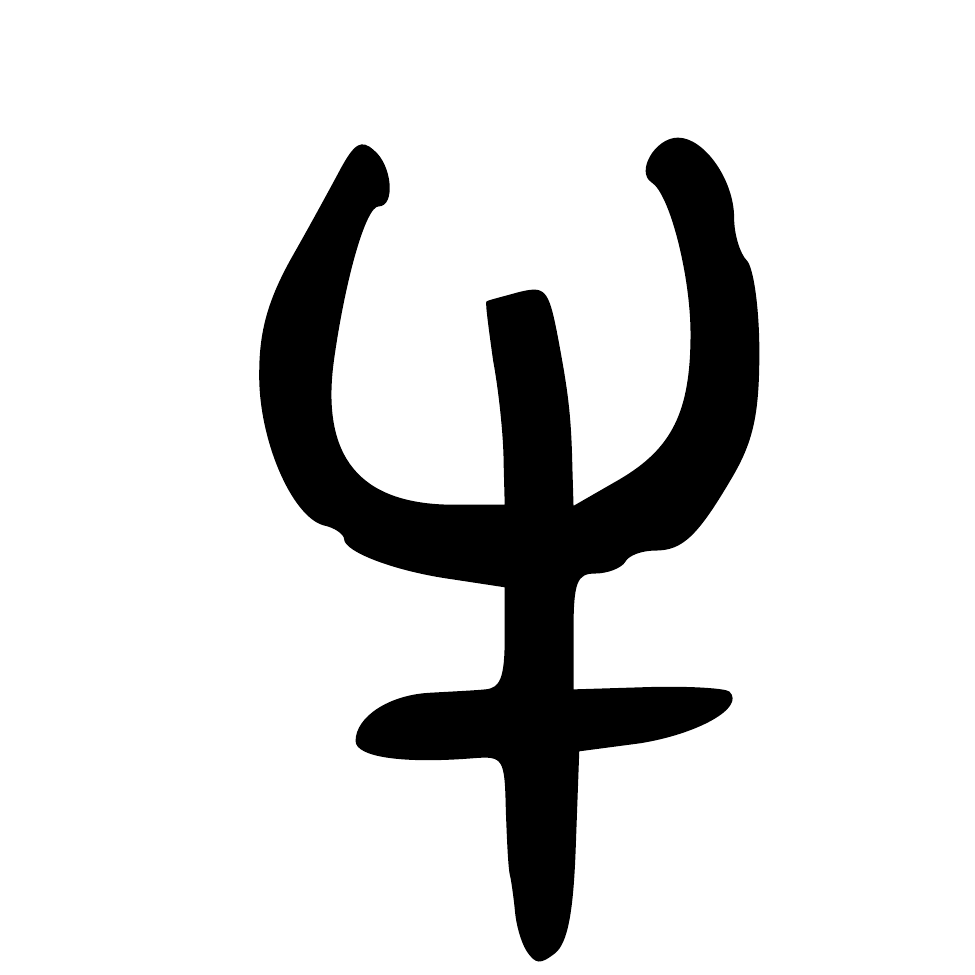} & $\rightarrow$ &
\includegraphics[height=1.15cm]{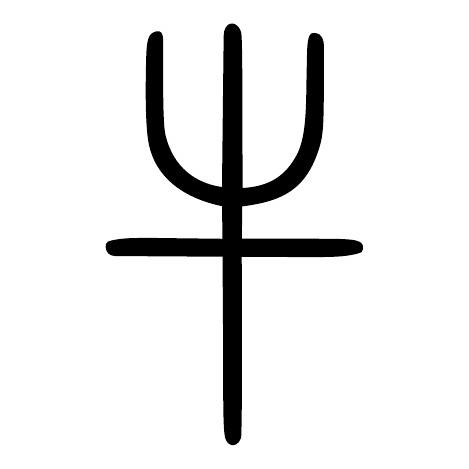} & $\rightarrow$ &
\includegraphics[height=1.15cm]{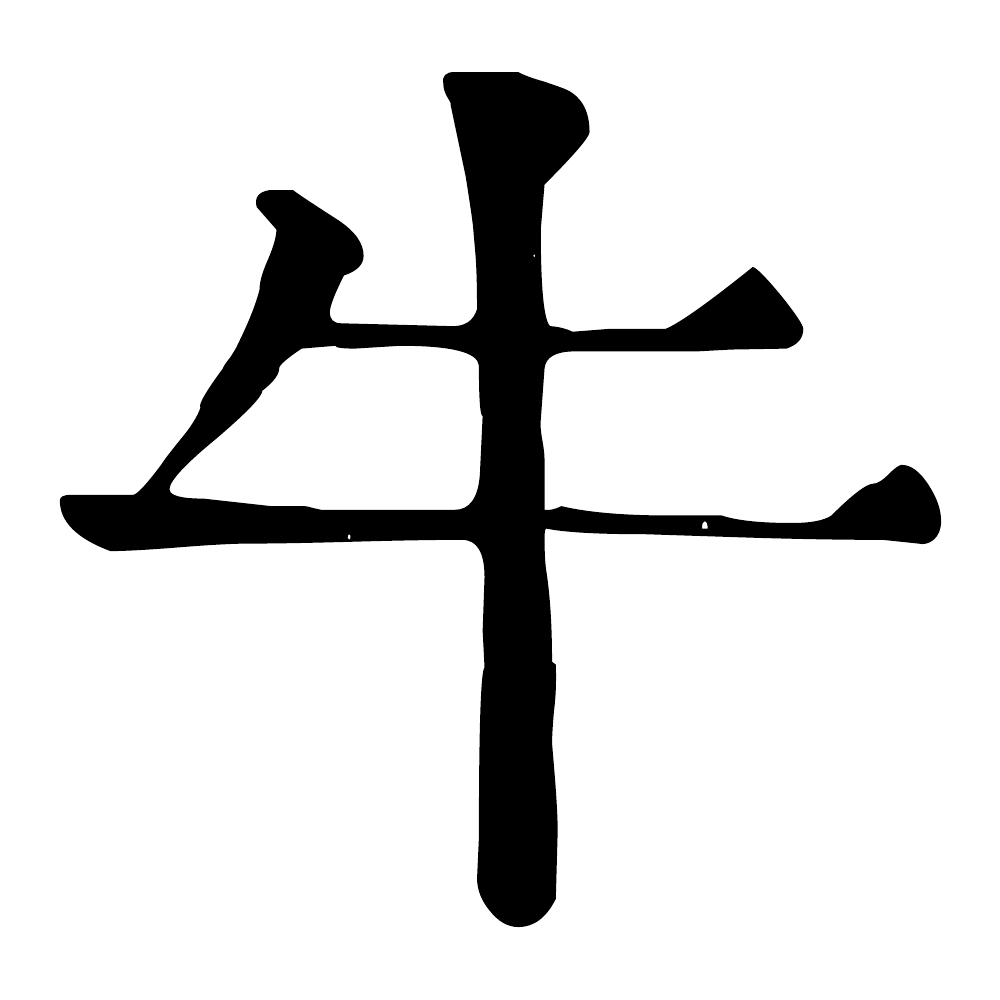}\\
\emph{shui} ``water'' &
\includegraphics[height=1.15cm]{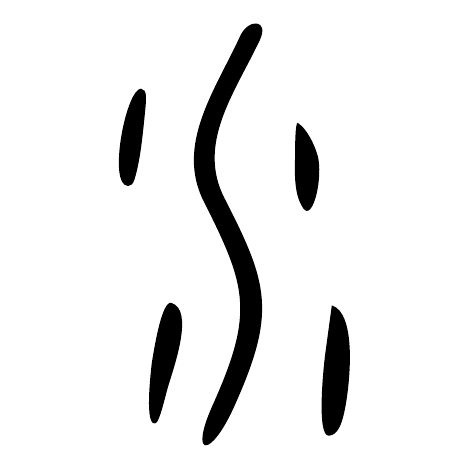} & $\rightarrow$ &
\includegraphics[height=1.15cm]{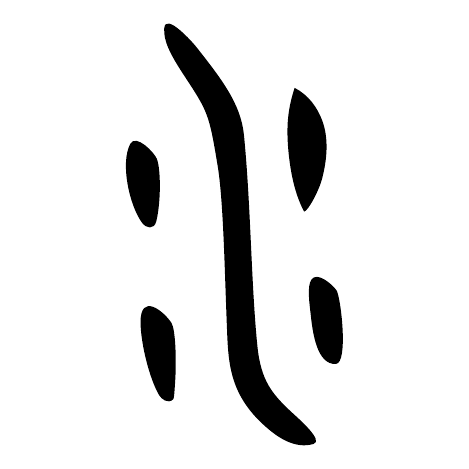} & $\rightarrow$ &
\includegraphics[height=1.15cm]{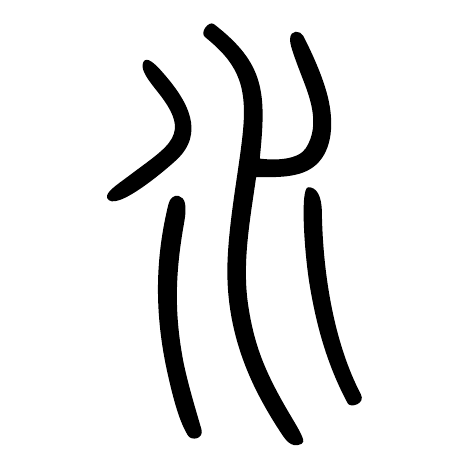} & $\rightarrow$ &
\includegraphics[height=1.15cm]{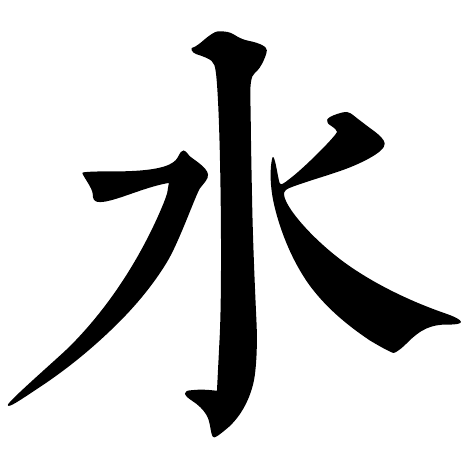}\\
& \footnotesize oracle bone & & \footnotesize bronze & & \footnotesize small seal & & \footnotesize modern
\end{tabular}
\caption{Two Chinese characters from oracle-bone pictograms (c.~1200 BCE) to the modern forms. Glyph drawings: Wikimedia Commons, public domain and CC licenses (see repository).}
\end{subfigure}
\caption{Three sequences of progressive abstraction. Each runs from a depiction to a sign that only a community can read.}
\label{fig:ladder}
\end{figure}

\begin{figure}[t]
\centering
\includegraphics[width=\linewidth]{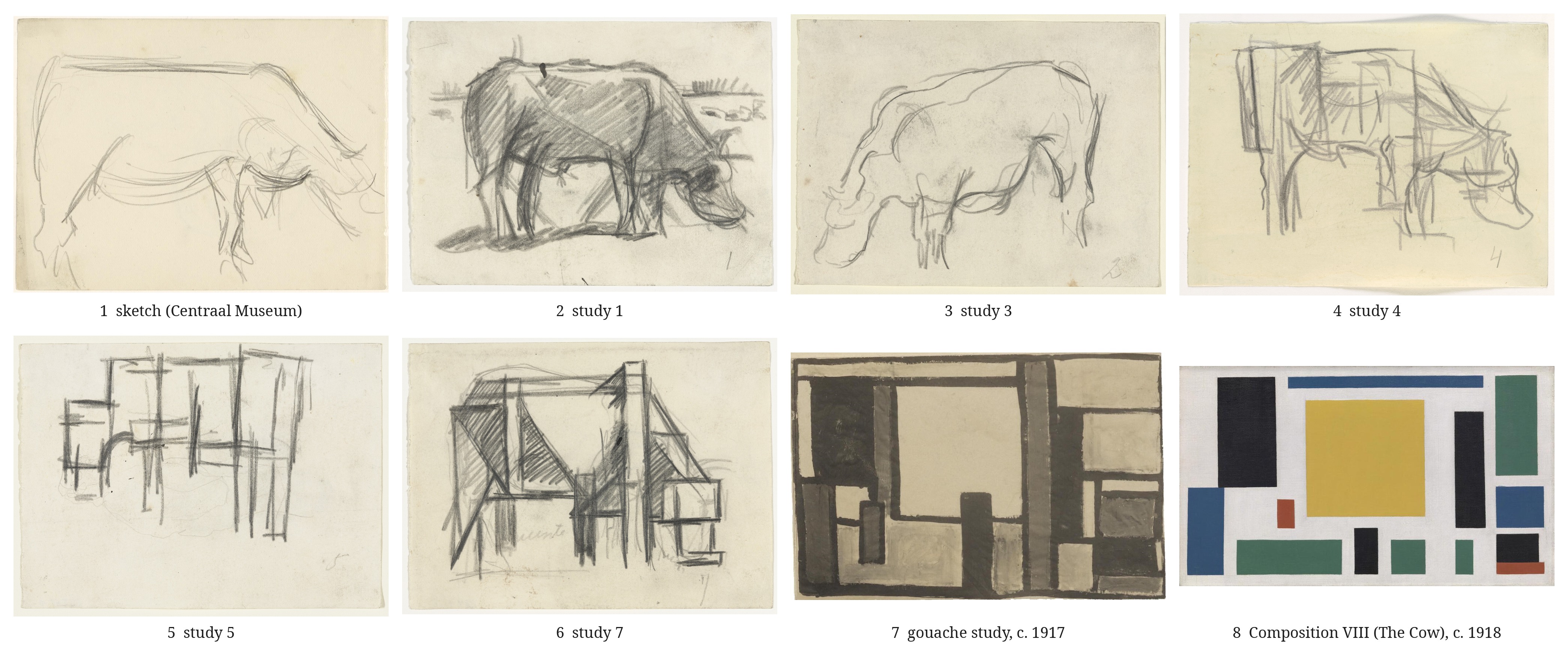}
\caption{Van Doesburg's cow series, 1917--18: eight states from a naturalistic sketch to a composition of fourteen rectangles. All works are in the public domain (Museum of Modern Art, New York, and Centraal Museum, Utrecht; images via Wikimedia Commons).}
\label{fig:cow}
\end{figure}

Picasso's \emph{Le Taureau} is a series of eleven lithographs made between December 1945 and January 1946. Each state removes lines from the one before, until a few strokes remain (Figure~\ref{fig:ladder}a). Van Doesburg's cow series of 1917--18 goes from a naturalistic sketch to a composition of fourteen rectangles (Figure~\ref{fig:cow}). Both series are sequences of progressive abstraction, and both stop one step short of the final abstraction, which is to replace the drawing with a word. The history of writing supplies that step. The letter A began as a pictogram of an ox head in the Proto-Sinaitic script. It became the Phoenician \emph{aleph}, named for the ox, was rotated and reinterpreted as the Greek alpha, and arrived in Latin as A \citep[][secs.~5, 21, and 24]{goldwasser2010,daniels1996} (Figure~\ref{fig:ladder}b). Chinese writing went through the same sequence. The oracle-bone form of \emph{niu} (ox) is a head with horns, the oracle-bone form of \emph{shui} (water) is a stream with drops, and both simplified into the modern characters (Figure~\ref{fig:ladder}c). Such pictographic origins are the exception, though. More than 80\% of the 9,353 characters in the \emph{Shuowen jiezi}, compiled around 100 CE, are phono-semantic compounds, and the share is higher in later dictionaries \citep[p.~84]{defrancis1984}.

The same sequence can be produced in the laboratory. In graphical communication experiments, drawings of a concept lose their pictorial detail over repeated exchanges and become arbitrary marks. The change requires feedback between partners, and people who did not take part cannot read the final signs \citep{garrod2007,fay2018}. Verbal descriptions shrink the same way \citep{clark1986}. CLIPasso takes the bull series as its model: it optimizes B\'ezier strokes against a CLIP-based loss and produces recognizable sketches at decreasing stroke counts \citep{vinker2022}. Image compression research now does the same thing. Text plus a sketch compresses images at very low bit rates \citep{lei2023}, and a caption plus a coarse latent, decoded by a diffusion model, reconstructs realistic images at 0.003 bits per pixel \citep{careil2024}. A language model trained primarily on text compresses ImageNet patches to 43.4\% of their size, against 58.5\% for PNG, and speech to 16.4\%, against 30.3\% for FLAC \citep{deletang2024}. Passing an image repeatedly through caption and regeneration shows how lossy the code is: some unified models drift within a few rounds, others hold \citep{mollah2025}.

The loss is measurable in human memory as well. Describing a face in words makes the describer worse at picking that face out of a lineup later, and the same holds for a color. The effect is called verbal overshadowing \citep{schooler1990}. A line drawing of two circles joined by a bar is redrawn from memory as eyeglasses or as a dumbbell, depending on which label was given at study \citep{carmichael1932}. Witnesses who were asked how fast the cars were going when they ``smashed'' reported higher speeds and, a week later, remembered broken glass that was not in the film \citep{loftus1974}.

Word categories also change perception while it is happening. Russian obligatorily distinguishes light blue (\emph{goluboy}) from dark blue (\emph{siniy}). Russian speakers discriminate two blues faster when they fall on opposite sides of that boundary than when they fall on the same side, and the advantage disappears under a verbal secondary task but not under a spatial one \citep{winawer2007}. \citet{gilbert2006} reported that this categorical perception is stronger in the right visual field, which projects to the language-dominant left hemisphere, a result that several later studies failed to reproduce (see below). Hearing an object's name can bring an image that has been made invisible by continuous flash suppression into awareness \citep{lupyan2013}. Saying the target's name aloud speeds up visual search, and the benefit turns into a cost as name and target diverge \citep{lupyan2012}. Infants at twelve months form categories more readily when the exemplars are labeled \citep{waxman1995}, and three-month-olds do so when the label is a word rather than a tone \citep{ferry2010}. The label-feedback hypothesis treats these results as a top-down bias on ongoing perception rather than a permanent rewrite of it \citep{lupyan2012label,lupyan2015}, and \citet{lupyan2016} describe the result as a mind that is ``programmable'' rather than merely trainable.

The oldest demonstration that a learned code intrudes on perception is the Stroop effect. Naming the ink color of the word ``red'' printed in green takes longer than naming the color of a plain square: in the original experiment, 110.3 against 63.3 seconds for 100 items, an increase of 74\%, while reading the words was unaffected by the colors they were printed in \citep{stroop1935}. The effect has held up through half a century of replications \citep{macleod1991}, and it depends on having learned the code. Children who cannot yet read show little interference, and it grows as reading becomes automatic \citep{schiller1966,macleod1991}. The word is irrelevant to the task and still wins, because the practiced pathway answers first. Section~\ref{sec:cueconflict} finds the same head start deciding which of two sufficient cues a robot policy uses.

Two of these effects have been tested at scale, with mixed results. A registered replication of the verbal overshadowing experiment found the effect at a fraction of its original size when the description immediately followed the video, and near its original size with the original timing. Across 31 laboratories, describing the face right after the video lowered correct identifications by 4 percentage points, against 22 to 25 points in the original experiments. With the timing of the original Experiment 1, a 20-minute delay before the description, 22 laboratories found a 16-point drop \citep{alogna2014}. The effect is real, and its size depends on timing. The lateralization of the color category effect has fared worse. Ten versions of the original experiments with 230 observers found category effects in both visual fields and no lateralization \citep{witzel2011}. A visual-search study found no category effect at the green-blue boundary in either field \citep{brown2011}, and a direct replication of the Gilbert et al. task, one of nine visual-field-asymmetry replications by \citet{brederoo2019}, did not reproduce the lateralized effect. \citet{witzel2016} summarize the accumulated results as casting serious doubt on the existence of genuine lateralized category effects, and their own experiments find categorical facilitation that is not lateralized. We therefore rely on the category effect that \citet{winawer2007} obtained with a verbal-interference control, and not on its lateralization.

The anatomy is consistent with the behavior. Semantic maps derived from silent films and from spoken stories occupy adjacent territory, and along the border of visual cortex the two maps line up category by category into one continuous map \citep{popham2021}. Learning to read sharpens early visual responses to horizontal features, creates an area selective for written words, competes slightly with faces at that site, and extends changes into V1; most of these changes appear even when reading is learned in adulthood \citep{dehaene2010}. People born blind acquire knowledge of visual appearance through language. Their judgments of animal shape, texture, and size largely agree with those of sighted people, and color is the dimension on which the two groups differ most \citep{kim2019}. Blind and sighted adults also share the same causal understanding of why objects have the colors they do \citep{kim2021}. A distributional model trained on text recovers part of the same appearance knowledge \citep{lewis2019}, and \citet{wang2026} review the evidence that text-trained models learn visual knowledge by the same route. Single neurons in the human medial temporal lobe respond to a person's face, to her written name, and to her spoken name \citep{quiroga2005,quiroga2009}. Multimodal language models develop object concepts whose dimensions match human similarity judgments and line up with category-selective visual areas \citep{du2025}, and language models predict activity best in regions whose responses are consistent across sentences, word clouds, and images \citep{ryskina2025}.

\subsection{Audition: perceptual narrowing, inner speech, and paralinguistic information}
\label{sec:hearing}

Language does not depend on hearing. Profoundly deaf signers processing sign language activate the left inferior frontal cortex and the planum temporale, the same sites that hearing speakers use for speech \citep{petitto2000}. The influence of language on hearing therefore has two parts: what language does to the auditory system, and what the voice carries that the codebook drops.

During the first year, the auditory system is tuned to the native language, a process called perceptual narrowing. Infants can tell non-native phonetic contrasts apart at six months and lose most of that ability by twelve \citep{werker1984}. Japanese infants' discrimination of English /r/ and /l/ declines between the ages of 6 to 8 months and 10 to 12 months, while American infants' discrimination improves over the same interval \citep{kuhl2006}. The tuning lasts. Among conservatory students who began training at the same age, absolute pitch is several times more common among speakers of Mandarin than among speakers of English \citep{deutsch2006}. Vision also reaches into hearing: a face mouthing /ga/ dubbed with /ba/ is heard as /da/ \citep{mcgurk1976}.

Reading involves inner speech. Intracranial recordings show top-down activation of voice-selective temporal cortex during silent word reading \citep{perrone2012}. The inner voice has the reader's own accent. Northern and Southern English readers are tripped up at different rhymes in limericks, according to how each would pronounce the final word \citep{filik2011}, and readers simulate the voices of characters they know \citep{kurby2009}. Inner speech is not universal, though. Adults who report little inner speech do worse on verbal working memory and rhyme judgments and equally well on other tasks \citep{nedergaard2024}, which parallels the roughly 2\% of people who report no visual imagery \citep{zeman2015}. The sensory format of language inside the head differs from person to person.

The voice carries information that text drops. Dogs show a hemispheric bias for meaningful words, process intonation in a separate auditory region, and their reward regions respond only when the two agree \citep{andics2016}. Prosody, speaker identity, and emotion travel outside the 39 bits per second of segmental content \citep{coupe2019}. Speech tokenizers show the split: the semantic tokens of a speech language model carry little information about speaker identity \citep{borsos2023}, and the units of generative spoken language models discard most prosodic information unless prosody is modeled separately \citep{kharitonov2022}. Audio language models built on speech-recognition encoders go further and answer questions about the voice from the transcript (Section~\ref{sec:ignore}). Naming a feeling changes the feeling: labeling the affect in an image lowers amygdala activity and raises activity in right ventrolateral prefrontal cortex \citep{lieberman2007}.

\subsection{Other modalities: olfaction, time, number, and space}
\label{sec:other}

Which senses a language can describe is a property of that language. Across 20 languages and five senses there are 13 different rankings of codability. Shape is the easiest domain to name in Malay and smell the hardest, Umpila shows the reverse, and in most languages smell is coded worst \citep{majid2018}. Jahai, spoken by hunter-gatherers in Malaysia, has about a dozen abstract odor terms, and its speakers name odors as consistently as colors, whereas English speakers fall back on describing the source \citep{majid2014}. Language also shapes how other dimensions are judged. Speakers of Swedish, who talk about duration in terms of length, are misled by the length of a line when they estimate duration. Speakers of Spanish, who talk about duration in terms of quantity, are misled by the fill of a container. Bilinguals switch with the language of the session \citep{bylund2017}. Reading ``kick'', ``pick'', or ``lick'' activates the leg, hand, or face region of motor cortex \citep{hauk2004}. The direction of writing sets the direction of the mental number line: Canadians place small numbers on the left, Palestinians on the right, and Israelis, who read words right to left and digits left to right, show no reliable direction \citep{shaki2009}.

\section{Language as a cognitive technology, and its dissociation from thought}
\label{sec:technology}

This section turns from perception to thought. Number words, writing, notation, and the research paper are technologies built on language (Section~\ref{sec:tools}); neuropsychology shows that thought survives the loss of language (Section~\ref{sec:dissociation}); and other species show concepts without a shared codebook (Section~\ref{sec:animals}).

\subsection{Symbolic technologies: number words, writing, notation, and the research paper}
\label{sec:tools}

Words are tools for thought in the same sense that a slide rule is a tool for arithmetic \citep{clark1998}. The Pirah\~a have no exact number words. They match even large sets one to one without error, but they fail as soon as the quantity has to be held in memory, so number words are a technology for storing quantity through abstraction \citep{frank2008}. Adults who shadow speech while being disoriented behave like rats and young children: they cannot combine the geometry of a room with the color of a wall. A rhythm-clapping task leaves the ability intact \citep{hermervazquez1999}. Learners of Nicaraguan Sign Language develop false-belief understanding as their vocabulary for mental states grows \citep{pyers2009}.

Writing is a technology that restructures thought \citep{ong1982}, and it is too recent to have shaped the genome. The cortical sites it uses were recycled from vision, a process known as neuronal recycling \citep{dehaene2007,dehaene2010}, and the recycling depends on the script: reading impairment in Chinese children involves the left middle frontal gyrus rather than the temporoparietal sites found in readers of alphabets \citep{siok2004}.

Mathematical notation pushes compression to its limit. \citet{aksenov2026} model mathematics as a hierarchy of macros. A definition or a theorem gives a name to a substring, and every later use of the name compresses the text. In Lean's Mathlib, the fully expanded length of a definition grows exponentially with its depth of nesting, while the wrapped length stays roughly constant \citep{aksenov2026}. Readers pay for this compression in memory. Immediate memory holds about seven chunks \citep{miller1956}, and chess masters recall boards by chunks that novices lack \citep{chase1973}. The same chunking is what lets a listener keep up with speech under the Now-or-Never bottleneck \citep{christiansen2016}. Whitehead wrote in 1911 that a good notation, ``by relieving the brain of all unnecessary work,'' sets it free for harder problems \citep{whitehead1911}, and Iverson's Turing lecture presented notation as a tool of thought \citep{iverson1980}.

The research paper is the same compression applied to research. A paper turns a branching search into a linear story. It drops the failed experiments and the implementation detail, which is what a reader limited to a few tens of bits per second (Section~\ref{sec:bandwidth}) wants removed. Agent-native research artifacts replace the story with an executable package that keeps the exploration graph and the raw evidence. On PaperBench and RE-Bench, question-answering accuracy rises from 72.4\% to 93.7\%, and on PaperBench reproduction success rises from 57.4\% to 64.4\% \citep{liu2026}. At the human end, the chat interface is a narrow serial channel: sequential text forces the user to verbalize, which invites verbal overshadowing of visual patterns \citep{reddy2026}.

\subsection{Neuropsychological dissociation of language from thought}
\label{sec:dissociation}

In humans, language can be separated from thought. \citet{fedorenko2024} argue from neuroimaging and neuropsychology that language is a tool for communication rather than for thought. People with severe aphasia after damage to the left hemisphere still solve arithmetic and algebra problems and reason about other people's minds \citep{fedorenko2024,klessinger2007,varley2000}. Three men with large left perisylvian lesions and severe agrammatism kept their computational procedures intact \citep{varley2005}. Understanding computer code engages the multiple-demand system and engages the language system weakly or not at all \citep{ivanova2020}. The language network is one component of the mind, and the other components can work without it.

\subsection{Comparative evidence: conceptual representation without a shared codebook}
\label{sec:animals}

Concepts do not require words. Dogs show a frontal mismatch response 206 to 606 ms after seeing an object that does not match a spoken word \citep{boros2024}. The timing is comparable to the N400, the human brain response to a word that does not fit its context, and the effect does not depend on how many words the dog knows \citep{boros2024}. Rhesus monkeys learn to associate Arabic numerals with numerosities, and their prefrontal neurons become tuned to both the sign and the quantity \citep{diester2007}. Single neurons in the marmoset hippocampus respond to the face and to the voice of the same individual \citep{tyree2023}, which puts the cross-modal invariance of human concept cells \citep{quiroga2009} in a non-human brain. Pigeons learn the category ``contains a person'' from photographs and transfer it to new images \citep{herrnstein1964}.

Apes that are given symbols use them. Kanzi, a bonobo raised with a lexigram keyboard and spoken English, understood novel spoken sentences at the level of a two-year-old child \citep{savagerumbaugh1993} and, in three experiments, tracked pretend objects, such as juice poured between empty containers, in response to verbal prompts \citep{bastos2026}. Chimpanzees with a history of token training and no language training judged relations between relations at the level of the language-trained chimpanzee Sarah. What mattered was the experience of associating arbitrary tokens with abstract relations \citep{thompson1997}. Symbols change what an ape can think, which is the comparative counterpart of the label-feedback results in Section~\ref{sec:vision}. Non-human animals have concepts, cross-modal representations of individuals, and isolated codebook entries. What they lack is a codebook of community scale that combines by rule.

\section{Language in multimodal models and language models}
\label{sec:models}

This section reads the human evidence against models. Vision-language models neglect the image, audio language models neglect the voice, and vision-language-action models neglect the instruction, and all three failures trace to the fusion rule (Section~\ref{sec:ignore}), which we then measure directly with cue-conflict experiments (Section~\ref{sec:cueconflict}). Representation, control, and reasoning can all be learned without language (Section~\ref{sec:without}). Language models are also the best current models of the human language network, and the relation runs in both directions (Section~\ref{sec:network}).

\subsection{Modality neglect in vision-language and vision-language-action models}
\label{sec:ignore}

Diagnostic studies of multimodal models split into two groups that point in opposite directions: models that answer questions neglect the image or the voice, and models that act neglect the instruction (Table~\ref{tab:ignore}).

\begin{table}[t]
\centering\small
\begin{tabular}{@{}>{\raggedright\arraybackslash}p{2.6cm}>{\raggedright\arraybackslash}p{2.2cm}>{\raggedright\arraybackslash}p{5.4cm}>{\raggedright\arraybackslash}p{4.6cm}@{}}
\toprule
Setting & Modality neglected & Evidence & Reported mechanism \\
\midrule
Vision-language models (question answering) & Vision & Fail on overlapping circles, line intersections, and ring counts \citep{rahmanzadehgervi2024}; CLIP-blind pairs \citep{tong2024}; answers survive total blurring, late layers suppress vision \citep{zhou2026}; success tracks whether the target has a name \citep{shahgir2026}; text wins conflicts \citep{deng2025words}; RL shortcuts \citep{xu2026} & The language prior predicts the answer better than the pixels do; benchmarks reward it \\
\addlinespace
Audio language models (paralinguistic questions) & Voice & When the transcript contradicts the voice, 12 models score 6 to 31\% and agree with the transcript 64\% of the time \citep{pang2026voxparadox} & Speech-recognition encoders discard prosody in late layers; the information that survives inside the model goes unused \\
\addlinespace
Vision-language-action models (manipulation) & Language & Paraphrase costs 22 to 52 points \citep{kim2026}; 90\% to 0.0\% under position and task perturbation \citep{zhou2025}; instructions ignored at fixed layout \citep{hou2026}; counterfactual instructions \citep{fang2026vision}; camera change 95\% to below 30\%, language change far smaller \citep{fei2025}; information collapse \citep{lian2026}; modality collapse \citep{zhan2026} & The image predicts the instruction; conditional mutual information between instruction and action vanishes \\
\addlinespace
Human perception & Neither & Reliability-weighted fusion \citep{ernst2002,alais2004}; causal inference on source \citep{kording2007}; modality appropriateness \citep{welch1980} & Weights follow current reliability and inferred common cause, not predictability \\
\addlinespace
Cue-conflict experiments (Section~\ref{sec:cueconflict}) & The redundant cue, whichever it is & OpenVLA-OFT obeys swapped instructions at 97\% where only they identify the task and at 0\% where the layout does, $\pi_{0.5}$ at 97\% and 28\%; six vision-language models weight image and text in the order of reliability at 0.11 to 0.82 of the ideal slope & A cue that adds nothing in training is dropped by one recipe and kept at a costly weight by another; a redundant cue added in fine-tuning is not learned; surviving cues are reliability-weighted, below the ideal \\
\bottomrule
\end{tabular}
\caption{Which modality each class of system neglects, the mechanism each study reports, and what the cue-conflict experiments add.}
\label{tab:ignore}
\end{table}

Vision-language models used for question answering tend to ignore the image. They fail at deciding whether two circles overlap or how many rings an Olympic-style logo has \citep{rahmanzadehgervi2024}, and models built on CLIP answer basic questions wrongly about image pairs that CLIP cannot tell apart but any person can \citep{tong2024}. Progressive blurring exposes the mechanism: across twelve VQA benchmarks and three models, a large share of answers survives severe or even total obscuring of the image, later layers suppress visual information in favor of text priors, and the benchmarks reward this behavior \citep{zhou2026}. Success tracks whether the relevant entity has a name \citep{shahgir2026}, and during reinforcement learning with verifiable rewards, video-language models stop watching the video unless a penalty is applied before the shortcut forms \citep{xu2026}. When text and image disagree on a vision task, ten models side with the text, and corrupted text lowers accuracy sharply \citep{deng2025words}. The problem is old: VQA models answered from question priors without looking at the image, which is what motivated the changing-priors splits \citep{agrawal2018}.

Audio language models fail the same way, with the transcript in the role of the language prior. VoxParadox puts the words against the voice: a young adult's voice says ``I am an elderly adult'', a male voice says ``being female, I feel empowered'', three overlapping voices say ``there is only one person speaking'', in 2{,}000 synthesized clips across ten paralinguistic tasks, from age and emotion to pitch and speaker counting \citep{pang2026voxparadox}. Twelve models, open and closed, score between 6 and 31\% on the acoustic answer and pick the transcript's answer 64\% of the time. Layer-wise probing locates two losses. Encoders pretrained for speech recognition suppress pitch, volume, and speaker information in their deeper layers as the representation aligns with the words, and where the information does survive into the language model, probes can read it while the model's answer still follows the text, a behavior the authors call lexical shortcutting. The second loss is the one Section~\ref{sec:hearing} anticipates: the codebook drops what the voice carries, and a model trained to reach the codebook drops it too.

Vision-language-action policies, in contrast, tend to ignore the instruction. Paraphrasing the instruction costs seven configurations of five such models between 22 and 52 percentage points \citep{kim2026}, and models that reach 90\% on the standard LIBERO suite fall to 0.0\% when object positions or the task itself are perturbed, while corrupted instructions leave their outputs unchanged \citep{zhou2025}. Translating the instruction into another language cuts success by 30 to 50\%, with the sensitivity concentrated at particular steps of the episode \citep{dong2026language}. Policies ignore the meaning of the instruction when the tabletop layout is held fixed \citep{hou2026}, and counterfactual instructions under familiar layouts expose the same shortcut, which a language-unconditioned branch used as a contrast at inference removes \citep{fang2026vision}. In a robustness analysis along seven dimensions, modest changes of camera viewpoint or initial state cut success from 95\% to below 30\% \citep{fei2025}. Language perturbations produced the second-smallest drop of the seven, 25 points, and removing the instruction left one model's success on the object suite largely unchanged; the preprint version says the models ``tend to ignore language instructions completely'' \citep{fei2025}.

\citet{lian2026} call the cause information collapse. In goal-driven datasets the instruction can be predicted from the image, so the conditional mutual information between instruction and action vanishes, and the policy degenerates into a vision-only policy that fails out of distribution. A Bayesian decomposition that forces a language-conditioned branch gains 11.3\% on the out-of-distribution SimplerEnv benchmark \citep{lian2026}. \citet{zhan2026} reach the same diagnosis under the name modality collapse, in which, in their words, ``strong visual priors overwhelm sparse linguistic signals'' and policies overfit to particular phrasings. Their remedy pairs paraphrase augmentation at training time with the subtraction of a vision-only affordance prior at decoding time, and \citet{zhang2026flatness} keep instruction following through fine-tuning with a flatter loss surface, without added data or architectural change. Representation tracing confirms that $\pi_{0.5}$ and OpenVLA are strong at visually grounded trajectories and weak at fine-grained semantic following \citep{shi2026}. Two studies qualify the picture: general VLM competence does not predict control performance and the vision encoder is the bottleneck \citep{zhang2026vlm4vla}, and pretraining on language-action pairs without images reduces dependence on scene-specific visual shortcuts \citep{lin2026}.

The two failures suggest one mechanism. The brain weights each sense by how reliable it is on the current dimension (Section~\ref{sec:fusion}). Our reading of the diagnostic evidence is that multimodal networks instead weight modalities by how well each predicts the training target. In a late-fusion network trained jointly, the modalities provably compete, and only a subset gets learned \citep{huang2022}. Modalities overfit at different rates \citep{wang2020}, a dominant modality suppresses the weaker one unless the gradients are modulated \citep{peng2022}, and contrastive image and text embeddings sit in separate regions of their shared space \citep{liang2022}. In question answering the language prior predicts the answer better, so vision is dropped; in speech, the transcript predicts the label, so the voice is dropped. In manipulation data the image predicts the instruction, so language is dropped. This reading contains two separable hypotheses: that the models lack reliability weighting altogether, and that they drop a cue that is redundant in training rather than down-weighting it. Section~\ref{sec:cueconflict} tests both. Explicit reliability machinery exists on both sides, from uncertainty-weighted and quality-aware fusion \citep{han2021,zhang2023,xue2023,zhang2024survey} to Bayesian causal inference as a unifying account of multisensory perception \citep{shams2022}.

Architectures with an explicit language-like bottleneck exist: the consciousness prior selects and broadcasts a low-dimensional state that ``makes it natural to map conscious states to natural language'' \citep{bengio2017}, and a shared global workspace replaces pairwise interactions between modules with a channel of limited bandwidth \citep{goyal2022}. A trained language model already contains one: its verbalizable representations form a reportable, steerable workspace that many circuits read and write \citep{gurnee2026workspace}.

Whether a system uses a cue turns out to depend on what the cue adds in its training data and on which pathway already carries the information, not on whether the system acts or answers (Section~\ref{sec:cueconflict}). The one function for which language clearly matters in a robot policy is generalization to new instructions out of distribution \citep{lian2026}, which is a communicative function.

\subsection{Cue-conflict experiments on models}
\label{sec:cueconflict}

\begin{figure}[tbp]
\centering
\includegraphics[width=\linewidth]{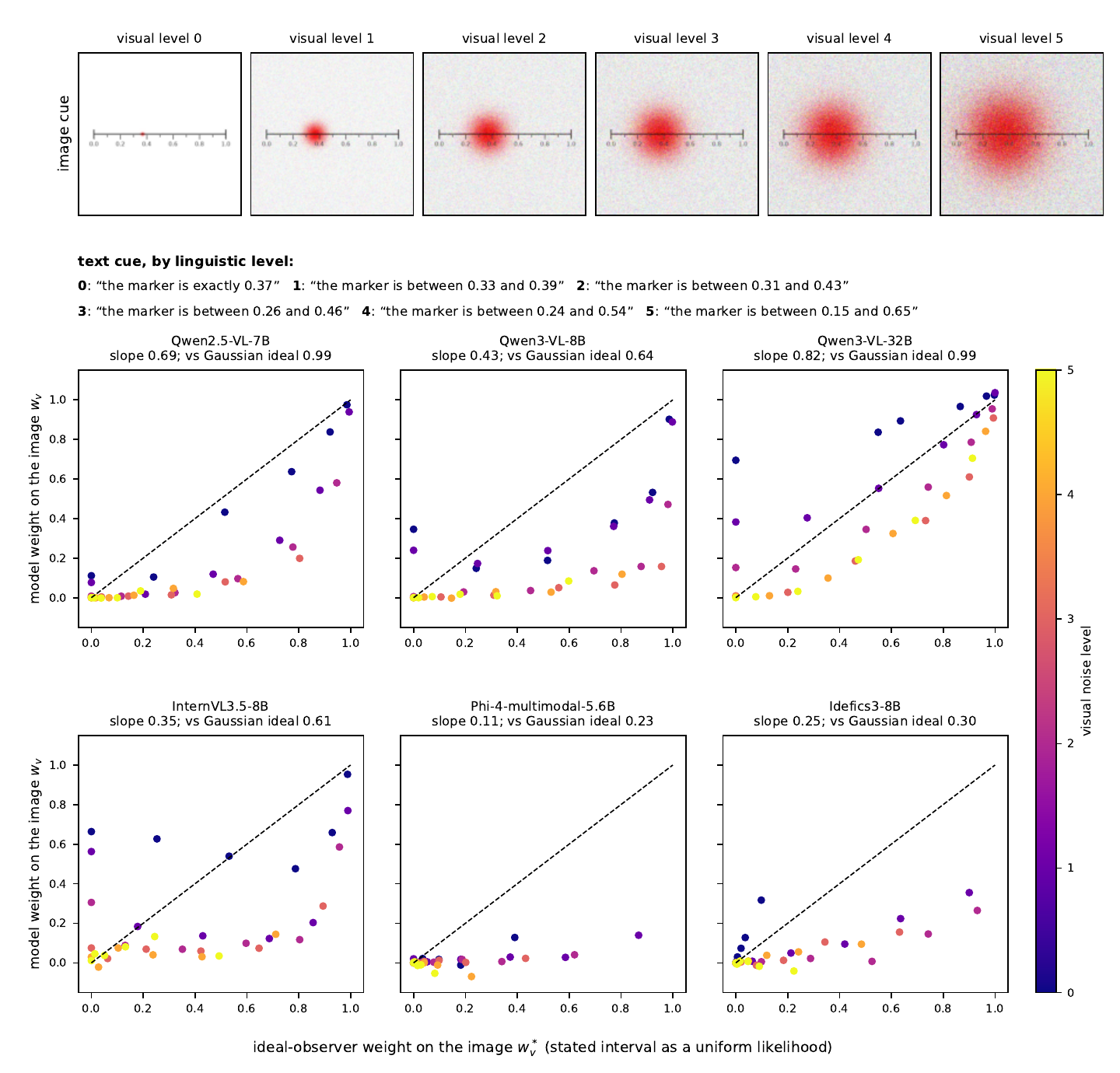}
\caption{Cue conflict in vision-language models. Top: the image cue, a marker drawn as a blob on a ruled line, at the six visual noise levels. Middle: the text cue at the six linguistic levels. Bottom: the weight each model gives the image on conflict trials, against the weight of an ideal observer that reads the image with the model's own precision and treats the stated interval as a uniform likelihood. Each point is one of the 36 reliability cells; the dashed line is the ideal observer; the slope against a Gaussian ideal is given for comparison.}
\label{fig:e1}
\end{figure}

The claims above rest on other groups' benchmarks. We also tested the fusion rule directly, with the cue-conflict method of psychophysics \citep{ernst2002}: give a model two cues to the same quantity, let them disagree slightly, and read each cue's weight off the answer. For each model we estimated the reliability of each cue from the model's own single-cue errors, derived the weights an ideal observer would use, and compared them with the observed weights. \citet{ma2025bayes} gave nine models an image and a text rendering of the same stimulus, blurred only the image, and found that many, though not all, of the models down-weighted it in a Bayes-consistent way. The experiments here vary the reliability of a natural-language cue as well as the visual one, manipulate what each cue predicts, and add two robot policies and a fine-tuning manipulation. Full protocols, item counts, and confidence intervals are in Appendix~\ref{app:methods}.

In the first experiment (Figure~\ref{fig:e1}) the model estimates the position of a marker on a ruled line from two cues. The image shows the marker as a blob whose width and pixel noise set the visual reliability, over six levels. A sentence states the position as an interval whose width sets the linguistic reliability, over six levels, from an exact value to a range of 0.5. On conflict trials the two cues disagree by 0.05 or 0.10, and the image value lies outside the stated interval on 39\% of them. The weight each model gives the image rises with the weight the ideal observer prescribes in all six models. Against a Gaussian ideal the slopes are 0.99 for Qwen2.5-VL-7B and Qwen3-VL-32B, 0.64 for Qwen3-VL-8B, 0.61 for InternVL3.5-8B, 0.30 for Idefics3-8B, and 0.23 for Phi-4-multimodal. Against an ideal that treats the interval as the uniform likelihood it is, the slopes fall to 0.69, 0.82, 0.43, 0.35, 0.25, and 0.11, with bootstrap intervals no wider than 0.13. The mean weight on the image is 0.01 to 0.49, against an ideal 0.15 to 0.54, and the two models that read the ruler worst, Idefics3-8B and Phi-4-multimodal, are also the two that fall furthest below the ideal. Two further facts qualify the picture. Between 39 and 78\% of the answers copy the stated interval's midpoint exactly, so a cell weight of 0.3 often means that the model copies the text on most trials and reads the image on the rest, the blind faith in text that \citet{deng2025words} report on vision-centric tasks. And the result depends on a legible image: with the ruler removed, the image error is 0.07 to 0.55 across noise levels and models, and five models give the image a mean weight of 0.00 to 0.07 where the ideal observer gives 0.02 to 0.20. Removing the instruction to combine the sources leaves the slopes at 0.71, 0.58, 0.52, 0.21, and 0.29 for the five models run without it. At the other extreme, a precise image is over-used: with a clean image and an exact sentence, Qwen3-VL-32B puts 0.70 of its weight on the image.

\begin{figure}[tbp]
\centering
\includegraphics[width=\linewidth]{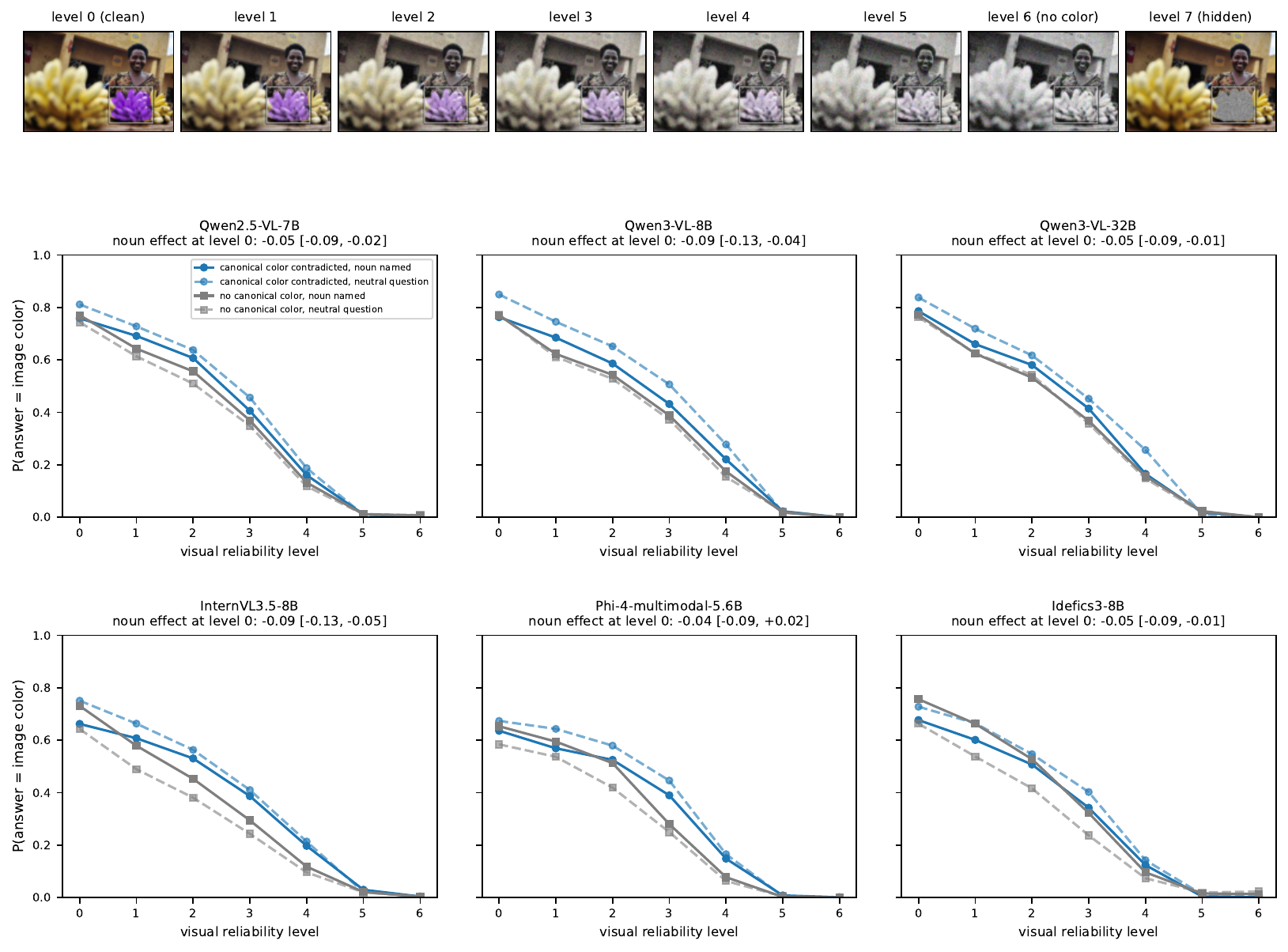}
\caption{The language prior against the image. Top: one stimulus series. A banana is recolored purple, the color of the whole photograph is drained over levels 1 to 6, and level 7 replaces the object with noise. Bottom: the probability that each model names the color shown in the image when the question names the object and when it does not, for nouns whose canonical color contradicts the image and for nouns without a canonical color, with the difference at the clean level and its bootstrap interval.}
\label{fig:e2}
\end{figure}

The second experiment sets the language prior against the image (Figure~\ref{fig:e2}). We recolored objects in COCO photographs to a color that contradicts the noun, a purple banana or a green stop sign, drained the color of the whole image over six levels beyond the clean image, and replaced the object with noise at a seventh. The design follows the visual counterfacts of \citet{golovanevsky2025pixels}, who find the prior dominant in early layers and the image in later ones. \citet{nooralahzadeh2026arbitration} show that the image is encoded correctly and the loss occurs when the model arbitrates between image and prior, and \citet{lietzow2026vision} localize the override to a few late attention heads. What we add is a graded axis of visual reliability and the contrast between a named and an unnamed object on the same image. Two questions per image differ only in whether the noun is named: ``What color is the banana inside the box?'' and ``What color is the object inside the box?''. Naming a noun with a canonical color lowers the probability of reporting the color in the image by 0.04 to 0.09 across the six models on clean images, with confidence intervals that exclude zero for five of them, and by 0.02 to 0.09 at the intermediate levels. For nouns without a canonical color the same manipulation changes the probability by 0.00 to 0.09 in the other direction. The image dominates: the models report the counterfactual color on 64 to 79\% of clean images. In relative terms the prior's pull grows as the color evidence weakens in five of the six models, from 6 to 12\% of the neutral answer at the clean level to 10 to 36\% at the fourth level; InternVL3.5-8B is the exception. A Bayesian prediction built from each model's own prior and likelihood failed a calibration check, so we report the contrast rather than a comparison with an optimum (Appendix~\ref{app:methods}).

\begin{figure}[tbp]
\centering
\includegraphics[width=\linewidth]{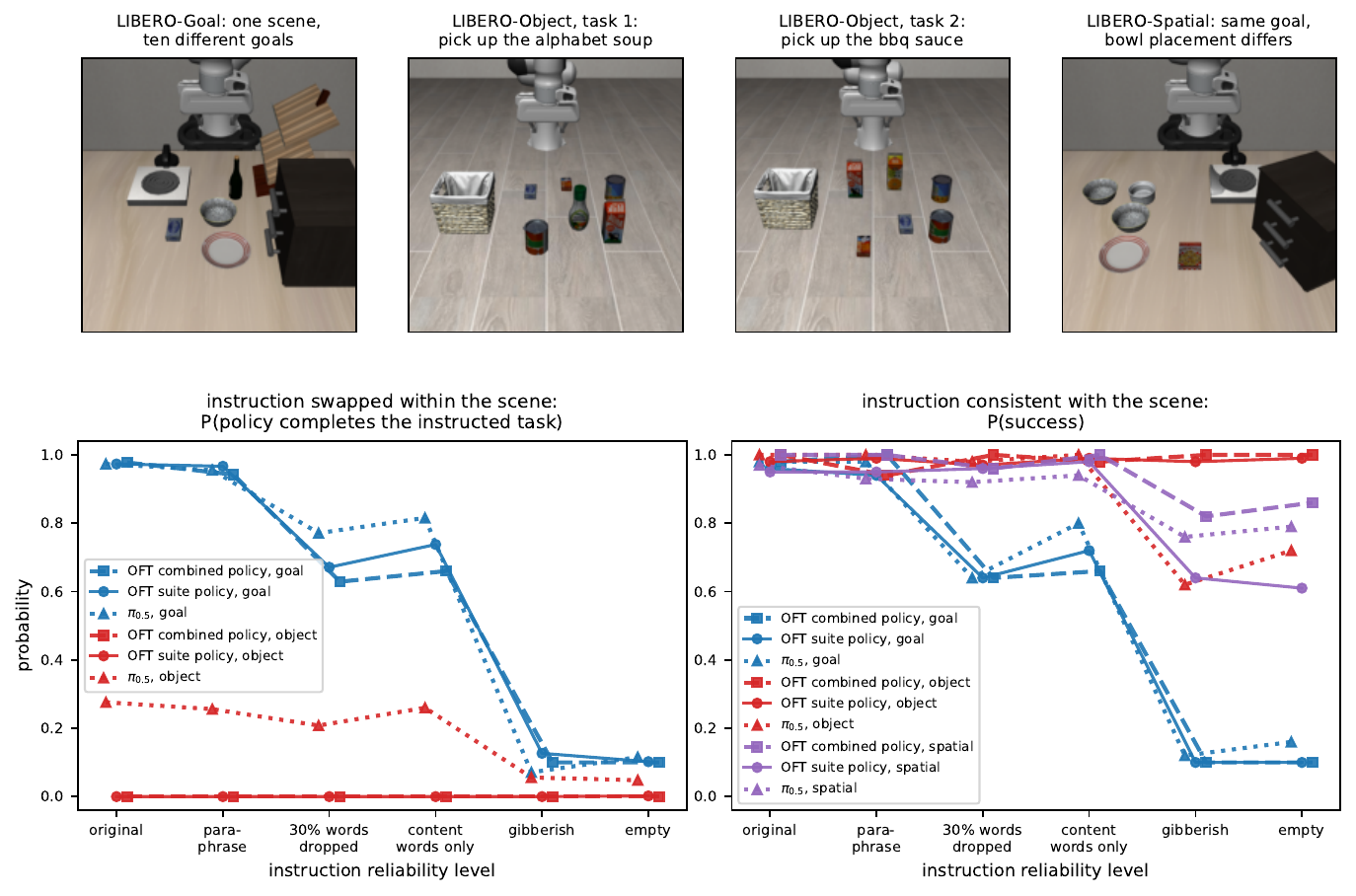}
\caption{Cue conflict in vision-language-action policies (OpenVLA-OFT and $\pi_{0.5}$ in LIBERO). Top: the suites differ in how much the instruction adds. One LIBERO-Goal scene supports ten goals; each LIBERO-Object task pairs its target with its own set of distractors; LIBERO-Spatial tasks share one goal and differ only in the initial placement. Bottom left: with the initial state fixed and the instruction swapped to another task of the same scene, the probability that the policy completes the instructed task. Bottom right: success with a consistent instruction as the instruction degrades. Solid lines are OpenVLA-OFT policies trained on one suite, dashed lines a single OpenVLA-OFT policy trained on all four, and dotted lines $\pi_{0.5}$; series are offset horizontally by 0.1 for visibility.}
\label{fig:e3}
\end{figure}

The third experiment asks the same question of a vision-language-action policy, OpenVLA-OFT \citep{kim2024openvla,kim2025oft}, in the LIBERO benchmark \citep{liu2023libero}, over 21{,}350 episodes, and then with a second policy family (Figure~\ref{fig:e3}). The three suites differ in what the instruction adds. In LIBERO-Goal one scene supports ten goals, so the instruction is the only cue to the task. In LIBERO-Object each task pairs its target with its own arrangement of distractors, so the layout predicts the task. In LIBERO-Spatial ten tasks share one goal and differ only in the initial placement, so the instruction is redundant by construction. We held the initial state fixed, swapped in the instruction of another task valid in the same scene, and degraded the instruction over six levels from the original sentence to gibberish and an empty string. We judged which task the policy completed by evaluating every task's goal predicates. The swapped-instruction design is that of LIBERO-CF \citep{fang2026vision}; we add the suite contrast, the attribution by goal predicates, and the graded degradation. Where the instruction is the only cue, the policy obeys it. On LIBERO-Goal it completes the instructed task on 97\% of swapped episodes with the original sentence or a paraphrase, on 74\% with content words only, and on 67\% under word dropout, with intervals of about 4 points on 450 episodes per level. With gibberish or an empty string it completes the instructed task on 10 to 13\%. Those completions are not residual instruction following: the policy then executes one default task, turning on the stove or putting the bowl on the stove, and the instructed task is completed only when it happens to be that task. Where the layout predicts the task, this policy ignores the instruction. On LIBERO-Object it completes the scene's own task on 97 to 100\% of swapped episodes and the instructed task on 0 to 1\% at every level, including original sentences that name another object in view, and gibberish leaves its success at 97 to 100\%. LIBERO-Spatial lies between, with 61 to 64\% success under unreadable instructions and an all-or-nothing pattern across tasks. A single policy trained on all four suites reproduces the ordering, with 10\% success on Goal, 82 to 86\% on Spatial, and 100\% on Object under unreadable instructions, so the checkpoint is not the explanation. Camera degradation is different in kind: on Goal it lowers the completion of the instructed task in the same proportion at every instruction level, from 97\% to 11\% for the original sentence, because a motor policy needs vision to act at all.

A second policy family shows that the wholesale dropping belongs to the training recipe rather than to manipulation as such. We ran the same protocol with $\pi_{0.5}$ \citep{intelligence2025pi05}, in its public LIBERO fine-tune, over 5{,}400 episodes. On LIBERO-Goal it reproduces the gradient: 97\% and 96\% for the original sentence and a paraphrase, 77\% and 82\% for word dropout and content words, and 7 to 12\% for gibberish and an empty string, again through one default task. On LIBERO-Object it does not drop the instruction. It completes the instructed task on 28\% of swapped episodes, the scene's own task on 30\%, and neither on 42\%, with the same split on every scene task. With unreadable instructions its success falls to 62 to 72\%, against 97 to 100\% for OpenVLA-OFT. On LIBERO-Spatial it keeps 76 to 79\%. So the two suites whose instructions add nothing behave alike under $\pi_{0.5}$, as the information account predicts, while OpenVLA-OFT treats them differently. The redundant cue is dropped by one training recipe and kept, at a weight that costs the policy two fifths of its episodes when the cues conflict, by the other.

\begin{figure}[tbp]
\centering
\includegraphics[width=\linewidth]{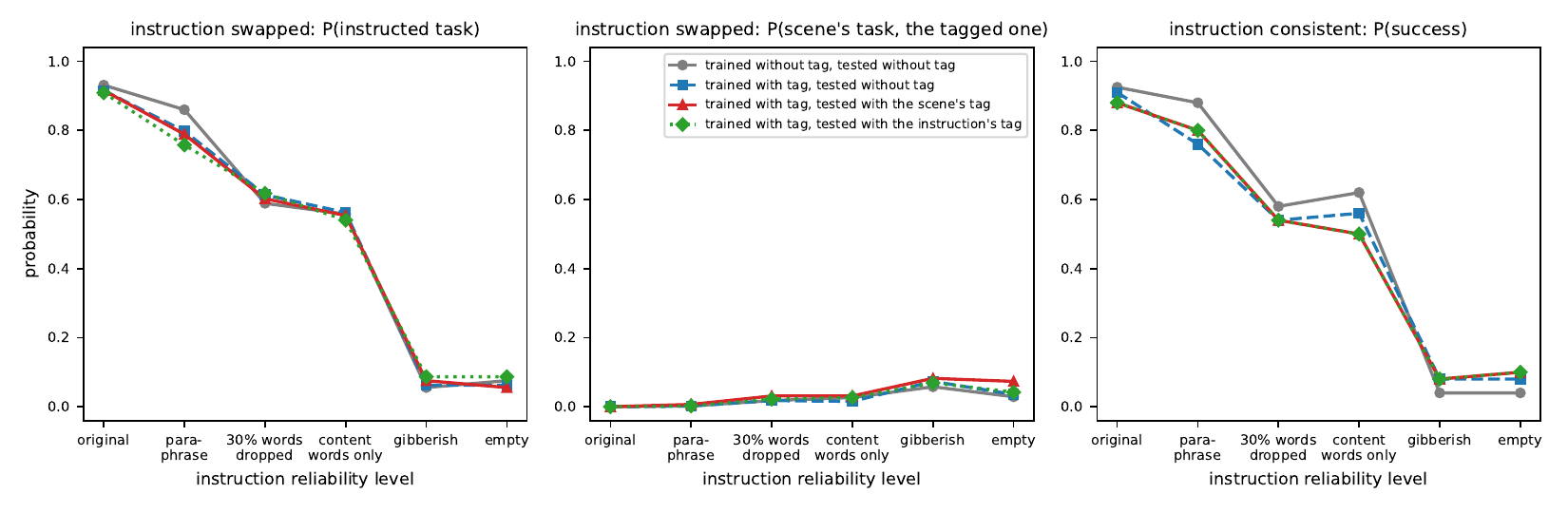}
\caption{Fine-tuning with a task-predictive visual tag on LIBERO-Goal. A colored square keyed to the task was painted on every training frame of one fine-tune and omitted from a control. Left: the probability that the policy completes a swapped instruction. Middle: the probability that it completes the scene's own task, the one the tag names. Right: success with a consistent instruction. Test conditions: the tagged policy with the scene's tag, without a tag, and with the instructed task's tag, and the control without a tag.}
\label{fig:e3t}
\end{figure}

The manipulation of the training distribution that would make this account causal did not behave as predicted. We fine-tuned the base OpenVLA model on the LIBERO-Goal demonstrations twice with the OFT recipe (Appendix~\ref{app:methods}). One run had a colored square painted in a corner of the third-person image, one of ten colors keyed to the task, so that the image identified the task in every training frame; the control run had no square. If predictability alone set the weights, the tagged policy should treat the instruction as the Object policies do. It did not. With the scene's tag present it still completed a conflicting instruction on 92\% of swapped episodes at the original sentence, against 93\% for the untagged control, and with the instruction unreadable it completed the tagged task on 7 to 8\%, against 3 to 6\% for the control. Removing the tag at test, or switching it to the instructed task, changed no level by more than 5 points (Figure~\ref{fig:e3t}). A cue that carried the whole task in every training frame was not learned in 15{,}005 steps of low-rank fine-tuning, because the instruction pathway, pretrained in the base model, already fit the data. This is the modality competition that \citet{huang2022} analyze for jointly trained networks: the pathway that reduces the loss first is learned and the other is not. The training statistics decide whether a cue is needed; a head start decides which of two sufficient cues wins, and in the Object suite the winning cue is the layout, which the vision encoder already represents. People show the same head start in the Stroop effect: a fluent reader cannot stop the word ``red'' printed in green from slowing the naming of the color, and a child who cannot read yet is unaffected (Section~\ref{sec:vision}).

Of the two hypotheses stated in Section~\ref{sec:ignore}, the first fails in part and the second depends on the system. Reliability weighting is present in all six vision-language models, in the direction the ideal observer prescribes but below its slope, and a coarse image is dropped rather than down-weighted. One policy family drops a redundant instruction entirely, and the other keeps it at a weight that the conflict then exposes. Across the three suites, the weight of the instruction follows the information it adds beyond the scene, which is the account of \citet{lian2026}. The ordering is a correlation over three suites, and the one manipulation of the training distribution we ran did not move it, because the added cue was never learned. This reading also reconciles the benchmark reports. Policies ignore corrupted instructions where the layout already predicts the task, as in the LIBERO-Object examples of \citet{zhou2025}, and fail to follow semantically new instructions in a fixed layout \citep{hou2026}, while the same architecture obeys in-distribution instructions where they carry the task. The suite dependence is visible in an earlier robustness study: with the instruction removed, the same policy keeps most of its success on LIBERO-Object and loses most of it on LIBERO-Goal \citep{fei2025}.

\subsection{Representation, control, and reasoning without linguistic supervision}
\label{sec:without}

Visual representations do not need language supervision. Visual self-supervised learning, scaled up in data and model size, matches or beats CLIP on visual question answering, including OCR and chart understanding, and it keeps improving at 7 billion parameters where CLIP saturates \citep{fan2025}. Control does not need a language backbone either. V-JEPA 2 pretrains on more than a million hours of video and post-trains an action-conditioned model on fewer than 62 hours of unlabeled robot video. It then picks and places objects zero-shot on Franka arms in two laboratories, with 65 to 80\% success depending on the object \citep{assran2025}. mimic-video replaces the vision-language backbone of a policy with a video generation model, on the argument that vision-language pretraining is blind to physical causality, and gains an order of magnitude in sample efficiency \citep{pai2025}. The position papers behind this line of work \citep{lecun2022,li2025} argue that much of what a model needs to know about the world is not in text at all. \citet{tenenbaum2026language} makes the cognitive-science version: children and animals think before language, so language is an interface to a model rather than the model.

Reasoning does not need language tokens. Coconut feeds the last hidden state back in as the next input. The resulting continuous thought encodes several candidate next steps at once, which lets the model search breadth-first where a verbal chain has to commit early \citep{hao2024}. Large Concept Models predict sentence embeddings autoregressively in a language-agnostic space \citep{lcm2024}. Visual Planning trains a vision model with reinforcement learning to plan in sequences of images and beats text-based planning on spatial tasks \citep{xu2025}; \citet{zhu2025} survey the field. \citet{gao2026reasoning} argue that reasoning after perception is reasoning in text, and their Turing Eye Test gives tasks that text cannot recover. One design already separates the two: the language model translates sentences into a probabilistic language of thought, and inference runs outside it \citep{wong2023word}. Inside language models, a small set of parameters is language-agnostic, and switching it off degrades performance in every language \citep{chen2025}, while the latent language of multilingual models leans toward English \citep{wendler2024}. This is the model-side counterpart of a human language network that is separate from reasoning \citep{fedorenko2024,ivanova2020}.

What is lost when reasoning leaves language is the ability to monitor it. A position paper from many organizations calls chain-of-thought monitorability a new and fragile opportunity for safety, and latent reasoning is one of the developments that would remove it \citep{korbak2025}. Agents that exchange hidden states instead of text argue that hidden states are more expressive than discrete tokens and that transferring them loses nothing \citep{zou2025}. A causal audit of latent multi-agent communication, however, finds that on one benchmark the aggregate effect decomposes into components of opposite sign, both of which reverse at a larger model scale \citep{zhang2026audit}. Two voice agents that switched from English to an acoustic data protocol on discovering that both were machines \citep{gibberlink2025} had left the human codebook. Section~\ref{sec:exit} returns to this trade-off.

\subsection{Language models as models of the human language network}
\label{sec:network}

Language models are the best available models of the human language network \citep{tuckute2024review}. Models that predict the next word better also predict neural responses to sentences better, and the strongest transformers explain close to 100\% of the explainable variance in fMRI and ECoG responses \citep{schrimpf2021}. Brains and autoregressive models share three computations: they predict the next word before it arrives, they register surprise when it does, and they represent words in context \citep{goldstein2022}. The brain also predicts further ahead, up to eight words, across a hierarchy of representations \citep{caucheteux2023}. A GPT-2 trained on 100 million words, about a decade of a child's input, already predicts brain responses \citep{hosseini2024}. Model embeddings track a word from the speaker's brain to the listener's \citep{zada2024} and place speakers of three languages in one conceptual space \citep{zada2025}. When the neuroscience localizer is applied to 18 language models, it finds language-selective units \citep{alkhamissi2025a}. Ablating them produces far larger deficits on language benchmarks than ablating random units, and they align more closely with the human language network than random units do \citep{alkhamissi2025a}.

The alignment has limits. Across 34 checkpoints and 300 billion tokens, brain alignment peaks between 2 and 8 billion tokens and then plateaus or declines, and it tracks formal linguistic competence rather than functional competence \citep{alkhamissi2025b}. That is the distinction \citet{mahowald2024} draw for language models: good at the rules and patterns of language, uneven at using language to get things done. At circuit level the separation is partial: formal mechanisms transfer among formal tasks better than to functional ones, without one unified language network \citep{hanna2026formal}. Surprisal from larger models fits reading times worse \citep{oh2023}. \citet{antonello2024} caution that brain-predictive features may be a by-product of next-word prediction rather than evidence for predictive coding. The limits are themselves informative. Language models model the language network, and the language network is not the whole mind.

The relation now runs in both directions, along four routes. The first is decoding. With a language model as a prior, a decoder reconstructs continuous language from fMRI, recovering the gist of heard speech, imagined speech, and silent films, and it works only with the subject's cooperation \citep{tang2023}. Intracortical speech neuroprostheses decode attempted speech with language-model priors at 62 words per minute on a 125,000-word vocabulary \citep{willett2023}, keep 97.5\% accuracy over 8.4 months \citep{card2024}, and decode inner speech, protected by a passphrase \citep{kunz2025}. A system with high-density surface electrodes decodes silently attempted speech into text at 78 words per minute with a 25\% word error rate, and also into synthesized voice and a talking avatar \citep{metzger2023}. The second route is stimulus design. Sentences that a GPT-based encoding model predicts will drive or suppress the language network do so in new participants, and surprisal and well-formedness are the main determinants \citep{tuckute2024}. The third route uses brain data as supervision. Fine-tuning speech models on fMRI responses to narratives improves their semantic representations and their downstream performance \citep{moussa2025}. Steering language models along directions shared between model and task-fMRI representations raises reasoning accuracy by up to 13 points across ten models from 1.5B to 72B parameters, and the gains are orthogonal to language-only supervision \citep{xiao2026}. This route was opened in 2019 by fine-tuning BERT to predict brain activity \citep{toneva2019}.

The fourth route is experiments that cannot be run on people: lesioning language units \citep{alkhamissi2025a}, controlling the developmental dose \citep{hosseini2024}, and testing theories of acquisition \citep{warstadt2022,piantadosi2023,warstadt2023}. A model trained on 61 hours of head-camera video and audio from a single child learns word-referent mappings and generalizes them \citep{vong2024}.

Two lessons follow for model builders: the language interface and the reasoning system can be separate components, as they are in the brain, and brain data are a supervision signal orthogonal to text.

\section{Language models and the human speech community}
\label{sec:exit}

This section treats language models as members of the speech community. Agents are leaving the human codebook for latent channels (Section~\ref{sec:latent}); model vocabulary is feeding back into human speech and writing (Section~\ref{sec:feedback}); and the distribution of words and concepts is narrowing as a result (Section~\ref{sec:narrowing}).

\subsection{Latent-channel communication between agents}
\label{sec:latent}

Agents have begun to leave the human codebook. In a 2025 demonstration, two agents that detected a machine on the other end switched to an acoustic protocol \citep{gibberlink2025}, and multi-agent systems now pass KV caches instead of text \citep{zou2025}. The gain is bandwidth: machines are free of the limit that holds human language at tens of bits per second \citep{zheng2024,coupe2019}. The cost is that a third party can no longer read the exchange, and being readable by bystanders is one of the services a shared codebook provides. End-task accuracy cannot say what a latent channel transmits; a causal audit splits the measured gain into the mere presence of a message and example-specific content, with components that flip sign across model sizes \citep{zhang2026audit}. The safety community asks that models keep reasoning in a language people can read \citep{korbak2025}. New codebooks for communities of agents are being written. Agent-native research artifacts replace the paper on the grounds that the paper was compressed for a bandwidth-limited human reader, and an agent is not so limited \citep{liu2026}. The test for such a codebook should be whether a person can audit it.

\subsection{Lexical feedback from models to speakers}
\label{sec:feedback}

The reverse process is further along and less discussed. Language models have joined the community that maintains the human codebook, and they are changing it. Words that ChatGPT overproduces, among them ``delve'', ``meticulous'', and ``intricacies'', rose abruptly after its release in about 740,000 hours of podcast episodes and in 360,000 academic talks, and a synthetic-control analysis attributes the rise to the model \citep{yakura2024}. A second analysis of 22 million words of unscripted podcast speech finds the same convergence \citep{anderson2025}. In writing, an analysis of excess vocabulary in more than 15 million PubMed abstracts implies that at least 13.5\% of 2024 abstracts were processed with a language model \citep{kobak2025}. Across 1,121,912 papers from January 2020 to September 2024, the estimated share of language-model-modified content reached 22\% in computer science and 9\% in mathematics and the Nature portfolio \citep{liang2024a}. An estimated 6.5 to 16.9\% of the text in peer reviews at AI conferences could have been substantially modified by a language model \citep{liang2024b}. The overuse is not explained by model architecture, algorithm choices such as tokenization, or training data, and the pattern is consistent with an origin in reinforcement learning from human feedback \citep{juzek2025}. A follow-up study on a Llama model ties the overuse to learning from human feedback more directly \citep{juzek2025b}. On this evidence, the change in the codebook points to alignment as its source. Speakers pick the words up because people already align their vocabulary with a machine partner, and do so more strongly when they believe the partner is a computer \citep{branigan2011}; they also shift register when addressing a voice assistant \citep{cohn2022}.

\subsection{Narrowing of the lexical and conceptual distribution}
\label{sec:narrowing}

The distribution is getting narrower. Writers who are given ideas by a language model produce stories that are rated more creative one by one and more similar as a set \citep{doshi2024}. Co-writing with an instruction-tuned model lowers lexical and content diversity \citep{padmakumar2024}, and users of ChatGPT as an ideation tool produce ideas that are less distinct from one another's \citep{anderson2024}. Aligned models show less conceptual diversity than their instruction-tuned bases and far less than human populations \citep{murthy2025}, in which a common noun already has an estimated 10 to 30 variants \citep{marti2023}. Training on recursively generated data removes the tails of the distribution, a process known as model collapse \citep{shumailov2024}. Models also hold covert prejudice against speakers of African American English that exceeds any human stereotype measured experimentally \citep{hofmann2024}. In Putnam's terms \citep{putnam1975}, part of the work of maintaining the codebook has passed to a system whose training objective compresses the distribution. \citet{yiu2024} and \citet{farrell2025} place large models among cultural technologies, alongside writing, print, and libraries, and \citet{brinkmann2023machine} treat machines as participants in cultural evolution. \citet{sourati2026homogenizing} review this evidence and conclude that shared reliance on a few models narrows how people write and reason. On the description used here, they are members of the language community.

Three feedback loops now run at once. The training distribution contains the model's own output \citep{shumailov2024}, alignment compresses conceptual diversity \citep{murthy2025}, and speakers converge on the model's vocabulary \citep{yakura2024,doshi2024}. If the diversity of the codebook is an asset of the community, it needs to be measured and protected as a design objective.

\section{Implications for token-based language models, and open problems}
\label{sec:claims}

We close with seven claims that the evidence supports, each paired with what it implies for a token-based system, and then list what remains open.

\begin{enumerate}
\item A word is an index, not content. Text alone yields the relational structure of a world model: Othello board states \citep{li2023}, linear coordinates of space and time \citep{gurnee2024}, and human-like object concepts, including in a text-only model \citep{du2025}. It yields the structure without the perceptual content \citep{harnad1990,lakemurphy2023}, which is why people born blind know the colors of animals worst \citep{kim2019}. Models learn that structure with a harder compression than people apply, recovering human category boundaries while discarding fine distinctions that people keep \citep{shani2026tokens}. How much meaning the structure alone carries is the open question of Section~\ref{sec:grounding} \citep{piantadosihill2022,sogaard2023}. Whatever lies outside the codebook, physical causality among other things, has to come in through perception and interaction \citep{bisk2020,fan2025,assran2025,pai2025}. Grounding does not require web-scale text; 61 hours of one child's experience are enough to teach word-referent mappings \citep{vong2024}.
\item Language is separable from thought in brains \citep{fedorenko2024}, and in models the formal competence of language is distinguishable from the functional competence of using it \citep{mahowald2024,alkhamissi2025b,hanna2026formal}. The separation is not yet built into the models: ablating their language-selective units disables every task posed in words \citep{alkhamissi2025a}. \citet{rothschild2025view} draws the opposite lesson from the same models: their inferential success shows how much of thought language itself can carry. The next step is a language interface designed separately from the reasoning system, which latent reasoning has begun to do \citep{hao2024,lcm2024,xu2025}.
\item Human language runs at tens of bits per second because the serial stage of human cognition does \citep{zheng2024,coupe2019}. The shared discrete codebook answers a different constraint: minds are separate, and the channel between them is noisy (Section~\ref{sec:bandwidth}). A model is bound by the second constraint whenever it talks to people, and by the first not at all. This justifies language at the model's interface and in the shared codebook, but not as the model's internal representation. For a token-based architecture, it puts the tokenizer at the boundary of the system and leaves the internal representation free of the human codebook.
\item Vision-language models neglect vision on question-answering benchmarks, audio language models answer from the transcript rather than the voice, and vision-language-action models neglect language in manipulation \citep{zhou2026,pang2026voxparadox,lian2026,huang2022}. Our cue-conflict experiments (Section~\ref{sec:cueconflict}) locate the rule as far as two policy families and one task family allow. A cue that is redundant in the training data is dropped rather than down-weighted by one policy family and kept at a weight that fails under conflict by the other. The cues that remain are weighted in the order their reliabilities prescribe but below the ideal observer's weight \citep{ernst2002,kording2007,ma2025bayes}. Neglect follows the task statistics and the head start of the pathway that already carries the information, not the modality as such: a visual tag that identified the task in every training frame was not learned while the instruction pathway already fit the data. The design target is therefore the wholesale dropping of coarse or redundant cues, and training distributions in which each channel keeps information the others lack, since adding a redundant cue does not by itself get it used.
\item Leaving the human codebook costs auditability \citep{zhang2026audit,korbak2025}, and the verbalizable workspace of \citet{gurnee2026workspace} is the part of a model that an audit can read. The arrangement is the brain's own: content stays in the sensory and motor systems, and only a small workspace is reported, through language \citep{baars1988,dehaene2001}. The report is not a faithful audit even in people, who explain their choices with reasons that did not cause them \citep{nisbett1977,johansson2005}, much as a chain of thought can omit what decided the answer \citep{turpin2023}. Auditability is therefore a matter of degree. A probe can read more than a model reports (Section~\ref{sec:ignore}), as brain decoding reads more than a person says (Section~\ref{sec:network}), and what has no name in the codebook is invisible to both, which ties this claim to claim 6. A new codebook for agents, such as the research artifact of \citet{liu2026}, should be judged by whether a person can audit it.
\item Language models have joined the human language community; they are shifting its word frequencies and, in co-writing studies, narrowing what people produce \citep{yakura2024,kobak2025,doshi2024,murthy2025,shumailov2024,sourati2026homogenizing}. Codebook diversity should become an explicit design objective.
\item Language models are the best current models of the human language network \citep{schrimpf2021,goldstein2022,hosseini2024,tuckute2024review}, and the relation can be used in reverse: for decoding \citep{tang2023,willett2023}, for stimulus design \citep{tuckute2024}, for brain data as supervision \citep{moussa2025,xiao2026}, and for experiments that cannot be run on people \citep{alkhamissi2025a,warstadt2022}.
\end{enumerate}

Several numbers in Section~\ref{sec:bandwidth} need replacing. The per-sense input rates come from Zimmermann's estimates of the 1980s \citep{zimmermann1989}, and there is no modern aggregate estimate for touch; a measurement of the information rate of the whole tactile channel would settle the figure. The cue-conflict experiments behind claim 4 cover one task family per system, and for manipulation two policy families in simulation; for speech we rely on the published conflict benchmark rather than our own runs. The ordering across suites is a correlation over three suites. The one training-distribution manipulation we ran, a task-predictive visual tag added in fine-tuning, was ignored by the policy, so the causal reading rests on the suite contrast and on the difference between the two policy families. The ideal-observer weights depend on the noise model assumed for the text cue, since a Gaussian and a bounded-uniform version differ in the fitted slopes by a factor of about 1.4. A Bayesian benchmark built from the models' own priors and likelihoods failed a calibration check on nouns with no color prior, so Section~\ref{sec:cueconflict} reports model-free contrasts. The measured weights shift with the prompt's wording, so the prompt is part of the stimulus, and four of the six vision-language models share a Qwen language backbone, while the two with other backbones read the ruler worst and fall furthest below the ideal. Several of the machine-learning results cited here appeared as 2026 preprints, and their numbers should be checked against the published versions. Codebook diversity, in claim 6, has no standard metric yet; the variant-counting method of \citet{marti2023} and the conceptual-diversity measures of \citet{murthy2025} are starting points.

The title asked where language should sit in a multimodal model, and the introduction listed four positions: input, output, internal representation, and the channel between agents. The seven claims sort them. Input and output are the human interface, where the codebook is shared with the user, and claims 1, 6, and 7 concern what a model takes from that codebook and does to it. The internal representation does not require language (claims 2 and 3), but what latent reasoning gives up is auditability (claim 5), so the choice is a trade between capacity and oversight rather than a matter of principle. The brain made the same trade: its content stays in sensory and motor systems, and only a small serial workspace is reported, through language, at tens of bits per second. Language at the boundary of a model is that division copied, not a compromise imposed on it. The channel between agents can leave the human codebook only if a person can still audit it (claim 5). And where a model already ignores one of its channels, claim 4 locates the cause in the training data and in the head start of the other channel, rather than in the modality.

\bibliographystyle{plainnat}
\bibliography{references}

\appendix
\section{Methods for the cue-conflict experiments}
\label{app:methods}

\paragraph{Models and decoding.} The vision-language models are Qwen2.5-VL-7B-Instruct, Qwen3-VL-8B-Instruct, Qwen3-VL-32B-Instruct, InternVL3.5-8B, Idefics3-8B-Llama3, and Phi-4-multimodal-instruct, all open weights, served with vLLM 0.24 at temperature 0. The Qwen models and Idefics3 answered within 24 tokens; Phi-4-multimodal ran with a 64-token limit; InternVL3.5-8B ignores a request for a bare number, so it ran with a one-sentence system prompt asking for the number alone and a 200-token limit, and we parsed the final stated estimate. The first four share a Qwen language backbone; Idefics3 uses a Llama 3.1 backbone and Phi-4-multimodal a Phi-4-mini backbone. The first robot policy is OpenVLA-OFT \citep{kim2025oft}, with the published LIBERO-Goal, LIBERO-Object, and LIBERO-Spatial checkpoints and the checkpoint trained on all four suites, run with the authors' evaluation code (two camera views, proprioception, chunks of 8 actions, 10 warm-up steps, the suite's step limit).

\paragraph{Experiment 1.} The image is a 448-pixel canvas with a horizontal line from 0 to 1, tick marks every 0.1 and labels every 0.2, and a red marker at a position drawn uniformly from 0.15 to 0.85. Visual level $v$ sets the marker's Gaussian blob width and the pixel noise: (3~px, 0), (20, 0.12), (36, 0.18), (48, 0.22), (60, 0.25), (80, 0.30). The text cue states the position as ``exactly $x$'' at level 0 or ``between $a$ and $b$'' with half-widths 0.03, 0.06, 0.10, 0.15, and 0.25 at levels 1 to 5, the interval centered within half its width of the true value. Image-only and text-only trials (200 per level) give each model's single-cue error. Bimodal trials (250 per cell, 36 cells) draw the text value at the image value plus $-0.10$, $-0.05$, 0, 0.05, or 0.10. The weight on the image is the coefficient of the image value in the regression of the answer on the image value and the interval midpoint; the Gaussian ideal weight is $\sigma_t^2/(\sigma_v^2+\sigma_t^2)$ from the single-cue mean squared errors, and the bounded-uniform ideal is the posterior mean of a Gaussian reading of the image, with the model's own $\sigma_v$, under the stated interval as a uniform prior, fitted with the same regression. Bootstrap intervals resample items within cells (200 replicates). The prompts state that both sources describe the same marker; a variant without the caveat that the text may be imprecise gave the reported numbers, and a variant that also omits the instruction to combine the sources gave slopes of 0.71, 0.58, 0.52, 0.21, and 0.29 for Qwen2.5-VL-7B, Qwen3-VL-8B, InternVL3.5-8B, Idefics3-8B, and Phi-4-multimodal against the bounded-uniform ideal. Without the ruler, image errors are 0.07 to 0.55 across models and noise levels, and the mean weights on the image are 0.00 to 0.07 against ideal weights of 0.02 to 0.20.

\paragraph{Experiment 2.} Objects come from COCO val2017 instance masks covering at least 5\% of the image, 25 per strong-prior noun (banana, broccoli, carrot, orange, stop sign, fire hydrant, elephant, sheep) and 12 per noun without a canonical color (car, bus, umbrella, suitcase, cup, bottle, chair, couch, kite, vase, surfboard, bench), 2{,}368 images in all. The masked object is recolored to a target hue that differs from its canonical color, or, for one fifth of the strong-prior items, to the canonical color as a control. Reliability levels scale the saturation of the whole image by 1.0, 0.6, 0.4, 0.25, 0.15, 0.08, and 0 with pixel noise up to 0.15, and level 7 replaces the object region with gray noise. A box marks the object. Answers are the first-token distribution over eleven color words. The Bayesian benchmark combined the answer to ``What color is a banana?'' (or the answer with the object hidden) with the answer to the neutral question as a likelihood; on nouns without a color prior, where the prior cannot matter, it mispredicted the observed probability by up to 0.32, and a likelihood measured on an interior texture patch did not remove the error, so Section~\ref{sec:cueconflict} reports the contrast between the two questions instead, with 134 to 139 items per level and bootstrap intervals.

\paragraph{Experiment 3.} Each LIBERO task supplies 50 initial states; we used the first 5 per condition for the Goal runs and the combined policy, and the first 10 for the Object and Spatial suite checkpoints. For a scene task $A$ and an instruction task $B$ valid in the same scene, the policy starts from $A$'s initial state and receives $B$'s instruction at one of six levels: the original sentence, a paraphrase, 30\% of words deleted, content words only, letters replaced at random, or an empty string. An episode ends when any task's goal predicates hold or at the suite's step limit, and the first task whose predicates hold is recorded. LIBERO-Goal has 9 valid swaps per scene (450 conflict episodes per level per checkpoint); LIBERO-Object 5 (250 for the combined policy, 500 for the suite checkpoint); LIBERO-Spatial none, because its ten tasks share one goal, so it contributes consistent-instruction runs only. Camera degradation applied Gaussian blur of 2, 4, and 8 pixels with pixel noise of 0.05, 0.10, and 0.15 to both camera images. The paraphrase level of LIBERO-Spatial was rerun with hand-written paraphrases and 10 initial states for both checkpoints, replacing a first version that only prefixed the original sentence. In total 21{,}350 episodes were run with OpenVLA-OFT, at about 6 seconds each. The second policy is $\pi_{0.5}$ \citep{intelligence2025pi05} in the public LeRobot LIBERO fine-tune (pi05\_libero\_finetuned), run with the LeRobot inference code under the same protocol with 5 initial states per task in all three suites: 3{,}000 Goal, 1{,}800 Object, and 600 Spatial episodes, 5{,}400 in all. For the training-distribution manipulation we fine-tuned OpenVLA-7B on the LIBERO-Goal demonstrations (428 episodes, 52{,}042 frames) with the OFT recipe, LoRA rank 32, 15{,}005 steps, image augmentation on, on four H200 GPUs, twice: once with a square of 32 of 256 pixels on a side, inset 8 pixels from the top-left corner of the third-person image, in one of ten colors keyed to the instruction, painted before augmentation so that random cropping trims it in some frames but leaves its color visible, and once without it. The swap protocol then ran on LIBERO-Goal with 5 initial states, 9 swaps, and six levels, plus 15 further consistent-instruction episodes per task, under four conditions: the tagged policy with the scene's tag, without a tag, and with the instructed task's tag, and the untagged policy without a tag, 3{,}150 episodes each. Wilson intervals on 450 episodes are within $\pm 0.05$.

\end{document}